\documentclass[11pt]{article}

\usepackage[final]{acl}

\usepackage{times}
\usepackage{latexsym}
\usepackage[size=tiny]{todonotes}

\usepackage[T1]{fontenc}

\usepackage[utf8]{inputenc}

\usepackage{microtype}

\usepackage{inconsolata}

\usepackage{graphicx}
\usepackage{subcaption}
\usepackage{booktabs}
\usepackage{multirow}
\usepackage{amsmath}
\usepackage{amssymb}
\usepackage{xcolor}
\usepackage{inconsolata}
\usepackage{pifont}
\usepackage{tabularx}
\usepackage{comment}

\usepackage{tikz}
\usetikzlibrary{shapes.geometric,arrows.meta,positioning,calc,fit,backgrounds,decorations.pathreplacing}

\newcommand{\chronolens}[1]{\textsc{ChronoLens}}

\tikzset{
  proc/.style   = {rectangle, rounded corners=2pt, draw=black!70, fill=blue!6,
                    text width=#1, align=center, inner sep=4pt, font=\small\sffamily, minimum height=8mm},
  proc/.default = 22mm,
  data/.style   = {rectangle, draw=black!70, fill=green!8,
                   text width=24mm, align=center, inner sep=4pt, font=\small\sffamily, minimum height=8mm},
  store/.style  = {cylinder, shape border rotate=90, aspect=0.25, draw=black!70, fill=green!10,
                   text width=24mm, align=center, inner sep=4pt, font=\footnotesize\sffamily},
  llm/.style    = {rectangle, rounded corners=2pt, draw=black!70, fill=orange!12,
                   text width=#1, align=center, inner sep=4pt, font=\small\sffamily, minimum height=8mm},
  llm/.default  = 22mm,
  human/.style  = {rectangle, rounded corners=2pt, draw=red!60!black, very thick, fill=red!6,
                   text width=#1, align=center, inner sep=4pt, font=\small\sffamily, minimum height=8mm},
  human/.default= 22mm,
  flow/.style   = {-{Stealth[length=2mm]}, draw=black!65, thick},
}

\title{\chronolens{}: Measuring Language Change Across \\
Time, Languages, and Linguistic Levels}

\author{
  \textbf{Gagan Bhatia}\textsuperscript{1},
  \textbf{Julian Schlenker}\textsuperscript{2}, \\
  \textbf{Simone Paolo Ponzetto}\textsuperscript{2},
  \textbf{Steffen Eger}\textsuperscript{1} \\
  \textsuperscript{1}University of Technology Nuremberg 
  \textsuperscript{2}University of Mannheim \\
  \texttt{(gagan.bhatia, steffen.eger)@utn.de}
}

\begin{document}
\maketitle
\begin{abstract}

Historical language change affects morphology, syntax, semantics, and pragmatics, yet computational studies typically examine these levels with incompatible representations and therefore cannot determine whether they evolve together across languages.
We address this problem by asking how the magnitude and direction of change vary across linguistic levels, languages, and historical periods within a single analytical space.
We introduce \textsc{ChronoLens}, a framework that combines frozen multilingual language models, feature-aligned crosscoders, and post-hoc linguistic interventions, and apply it to 44.98 million documents and approximately 17.2 billion tokens from five parliamentary traditions spanning 1803--2026.
The resulting sparse representations agree substantially more strongly with linguistic statistics than dense embeddings or a pooled sparse autoencoder ($\rho=0.72$ versus $0.29$ and $0.28$), and reveal that morphology, syntax, semantics, and pragmatics generally change by comparable amounts within a language, while languages differ markedly in when, how far, and in which direction they change.
These findings show that historical language change is a structured, multidimensional process: similar magnitudes can conceal different trajectories, and meaningful cross-linguistic comparison requires measuring both distance and direction. \footnote{Code and data are available at \url{https://github.com/NL2G/Chronolens}.}

\end{abstract}

\section{Introduction}
\label{sec:introduction}

Language change over time affects morphological form, syntactic structure, meaning, and
pragmatic function, but computational studies rarely examine these levels
within the same analytical framework
\citep{degaetano-ortlieb-teich-2018-using,
bizzoni-etal-2019-grammar,reinig-etal-2024-politics}.
Most computational research on historical language change has focused on
lexical semantics, using static embeddings, contextual representations,
usage similarity, or optimal transport to compare word meanings across
periods
\citep{hamilton2016diachronic,giulianelli-etal-2020-analysing,
periti-tahmasebi-2024-systematic,periti-montanelli-2024-survey,
kishino-etal-2025-quantifying}.
Morphological comparisons commonly use annotated form distributions, while
computational studies of pragmatics often model speech acts or other
communicative functions
\citep{berdicevskis-etal-2018-using,baayen2009morphological,
reinig-etal-2024-politics,subramanian-etal-2019-target}.
Because these research areas use different representations, datasets,
and units of measurement, their trajectories cannot be compared directly
\citep{degaetano-ortlieb-teich-2018-using,
bizzoni-etal-2019-grammar,periti-montanelli-2024-survey}.
This incompatibility prevents us from answering a basic question about
language change: do morphology, syntax, semantics, and pragmatics change
together, or do they follow distinct historical trajectories?
Studies relating lexical and grammatical development suggest that changes at
different levels can interact, but these analyses have largely focused on one
language, register, or pair of linguistic measurements
\citep{degaetano-ortlieb-teich-2018-using,
bizzoni-etal-2019-grammar,chen-etal-2026-syntactic}.
A change observed in one language 
may reflect a broader development shared by
several linguistic communities, or it may result from that language's
grammar, political history, or corpus composition
\citep{hamilton-etal-2016-cultural,niu-etal-2023-cross,
krielke-etal-2024-crosslinguistic}.
Determining which explanation is more plausible requires a representation in
which linguistic levels, languages, and historical periods are directly
comparable.

Multilingual language models provide part of this representation because they
encode several languages within a common dense space
\citep{schuster2019crosslingualalignmentcontextualword,
martins2024eurollm}.
Dense model dimensions, however, do not correspond to stable linguistic
features because linguistic information is distributed across dimensions and
individual dimensions may participate in several unrelated computations
\citep{bricken2023monosemanticity,cunningham2023sparse,
templeton2024scaling}.
Sparse autoencoders address this problem by decomposing dense activations into
a sparse set of more selective features
\citep{cunningham2023sparse,gao2025scaling,
lieberum2024gemmascope}.
Yet sparse autoencoder features are not canonical: separate training runs or
datasets need not recover the same feature inventory
\citep{leask2025canonical,karvonen2025saebench}.
Consequently, feature $j$ in a dictionary trained on one language or period
has no guaranteed correspondence to feature $j$ in another dictionary
\citep{leask2025canonical,deng-etal-2025-unveiling}.
Recent approaches 
connect either languages and features, corpora and time, or
model checkpoints and linguistic capabilities.
To our knowledge, no prior work places multiple human languages, historical
periods, and linguistic levels in a shared, feature-aligned representation
framework
\citep{deng-etal-2025-unveiling,jing-etal-2026-histlens,
bayazit-etal-2026-crosscoding,fedorova-etal-2026-dhplt}.

We introduce \chronolens{}, a framework for comparing historical change across five languages and four linguistic levels within a common analytical space. We study English, German, Italian, Polish, and Turkish using a unified corpus
of 44.98 million parliamentary documents and approximately 17.2 billion tokens collected from 22 open sources spanning 200+ years
\citep{coole-etal-2020-unfinished,blatte-blessing-2018-germaparl,
ogrodniczuk-niton-2020-new,Cova_2025,Gngr2018ACO,
erjavec2023parlamint,Erjavec2024}. 
Parliamentary proceedings provide dated records produced under recurring institutional roles and communicative conventions, which has made them a common resource for comparative political and historical language research
\citep{erjavec2023parlamint,Erjavec2024,
de-jong-etal-2024-parlamint,skubic-fiser-2024-parliamentary}. Their long, precisely dated coverage and relatively stable institutional context facilitate cross-temporal and cross-lingual comparison while limiting variation due to changing genres.
Previous parliamentary studies have examined ideological change, political framing, solidarity, and speech acts, but generally within one language or with one predefined linguistic outcome
\citep{walter2021diachronicanalysisgermanparliamentary,
kostikova-etal-2024-fine,ghafouri-etal-2025-framing,
reinig-etal-2024-politics}.

Our contributions are as follows:
\textbf{(i)} we introduce \textsc{ChronoLens}, a unified framework for comparing historical language change across five languages, multiple historical periods, and four linguistic levels: morphology, syntax, semantics, and pragmatics;
\textbf{(ii)} we construct a multilingual diachronic corpus comprising 44.98 million parliamentary and political documents and approximately 17.2 billion tokens from 22 open sources, spanning the period from 1803 to 2026;
\textbf{(iii)} we develop a feature-aligned methodology that combines frozen multilingual language models with crosscoders and post-hoc probe interventions, enabling sparse features to be compared directly across languages and periods without using linguistic labels during feature learning; and
\textbf{(iv)} we provide an empirical analysis showing that the resulting features are more strongly aligned with observed linguistic statistics than dense embeddings or sparse autoencoders, and that historical change is coordinated across linguistic levels but differs substantially across languages in its timing, magnitude, and direction.

\section{Related Work}
\label{sec:related-work}

\noindent\textbf{Computational approaches to language change.}
Most computational work on diachronic change focuses on lexical semantics
\citep{hamilton2016diachronic,eger-mehler-2016-linearity,
periti-tahmasebi-2024-systematic,kishino-etal-2025-quantifying}, while
syntactic change is commonly measured through dependency distance and
structural complexity
\citep{liu-etal-2022-dependency,krielke-etal-2025-tracing,
chen-etal-2026-syntactic}. Work on morphology has examined productivity,
morphosyntactic complexity, and the relation between morphological structure
and meaning
\citep{baayen2009morphological,berdicevskis-etal-2018-using,
cotterell-schutze-2018-joint,nagata-etal-2026-cross}. Moreover, shared
processing pressures such as dependency-length minimization provide a reason
to expect partial convergence across languages
\citep{futrell-etal-2015-large,gibson-etal-2019-efficiency,
futrell-etal-2020-dependency,hahn-xu-2022-crosslinguistic,
niu-etal-2023-cross,xu-futrell-2024-syntactic}. However, prior studies
generally apply different representations to different phenomena. We instead
compare morphology, syntax, semantics, and pragmatics within one shared
representation space.

\noindent\textbf{Parliamentary and political discourse.}
Parliamentary corpora have supported diachronic and comparative research on ideology, framing, migration, solidarity, and speech acts
\citep{erjavec2023parlamint,Erjavec2024,
walter2021diachronicanalysisgermanparliamentary,
ghafouri-etal-2025-framing,kostikova-etal-2024-fine,
reinig-etal-2024-politics}. This work typically targets one language, concept, or prediction task. In contrast, we use comparable parliamentary material to investigate whether several linguistic levels follow shared or language-specific historical trajectories.


\noindent\textbf{Sparse and feature-aligned representations.}
Sparse autoencoders recover interpretable, language-selective, and culturally
selective features from language-model activations
\citep{cunningham2023sparse,deng-etal-2025-unveiling,
andrylie2025sparse,zou-etal-2026-deciphering,jing-etal-2026-histlens}.
However, independently trained dictionaries need not contain aligned features
\citep{leask2025canonical}. Crosscoders address this problem by learning a
shared feature index across models or checkpoints
\citep{lindsey2024crosscoders,
jiralerspong2026crossarchitecturemodeldiffingcrosscoders,
minder2026overcomingsparsityartifactscrosscoders,
bayazit-etal-2026-crosscoding}. We adapt them to languages and historical
periods, assign linguistic interpretations post hoc, and distinguish the
magnitude and direction of change.
\begin{table}[t]
\centering
\small
\resizebox{\linewidth}{!}{
\begin{tabular}{llrrc}
\toprule
\textbf{Language} & \textbf{Code} & \textbf{Docs (M)} & \textbf{Tokens (B)} & \textbf{Coverage} \\
\midrule
English & en & 17.78 & 6.8 & 1803--2026 \\
Italian & it & 5.15  & 3.6 & 1848--2022 \\
German  & de & 4.45  & 3.6 & 1867--2026 \\
Polish  & pl & 15.61 & 2.5 & 1919--2025 \\
Turkish & tr & 1.99  & 0.7 & 1950--2023 \\
\midrule
\textbf{Total} & & \textbf{44.98} & \textbf{17.2} & \textbf{1803--2026} \\
\bottomrule
\end{tabular}
}
\caption{Corpus overview by language. The full corpus contains approximately 17.2B tokens.}
\label{tab:corpus-main}
\end{table}

\section{Dataset}
\label{sec:dataset}

\noindent\textbf{Sources and coverage.}
We introduce a multilingual diachronic corpus of parliamentary speech and related political text. The corpus draws on 22 open official and research sources across five languages, combining long parliamentary records with smaller complementary sources such as party manifestos. 
The main parliamentary sources include UK Hansard and TheyWorkForYou for English \citep{coole-etal-2020-unfinished,https://doi.org/10.5281/zenodo.591264}, the Polish Parliamentary Corpus and ParlaMint for Polish \citep{ogrodniczuk-niton-2020-new,Erjavec2024}, ItaParlCorpus, IPSA, and ParlaMint for Italian \citep{Cova_2025,frasnelli-palmero-aprosio-2024-theres,Erjavec2024}, GermaParl, German parliamentary proceedings, Reichstag material, DeuParl, ParlaMint, and official Bundestag records for German 
\citep{blatte-blessing-2018-germaparl,walter2021diachronicanalysisgermanparliamentary,Erjavec2024}, and TBMM and ParlaMint data for Turkish \citep{Gngr2018ACO,Erjavec2024}. Party manifestos are drawn from the Manifesto Project \citep{https://doi.org/10.25522/manifesto.mpds.2017b}. All sources are mapped to a unified record schema with shared fields for language, date, document type, source, and text. This schema is necessary for cross-lingual comparison, since a metric can only be compared across languages when the underlying records carry the same temporal and document-level metadata. 
The final corpus contains 44.98M documents and approximately 17.2B tokens, spanning 1803 to 2026. Table~\ref{tab:corpus-main} summarizes the corpus by language. Appendix~\ref{app:sources} lists the individual sources.

\noindent\textbf{Quality control and density.}
Because historical OCR error can look like language change, we apply source-specific quality control before sampling. For OCR-derived Reichstag material from 1867 to 1942, we use ABBYY FineReader character confidence and remove pages below a confidence threshold or with more than 15\% low-confidence glyphs, which removes about 8\% of the oldest pages. Born-digital sources do not have OCR confidence scores, so we screen them with a character-\(n\)-gram gibberish detector, which removes about 0.5\% of born-digital text. We treat pre-1949 data cautiously because OCR noise is concentrated there. Since diachronic analysis also requires continuous coverage, not only large total size, Figure~\ref{fig:coverage} reports both the temporal span of each language and per-decade document density after OCR filtering.

\section{\chronolens{}}
\label{sec:method}

\begin{figure*}[t]
    \centering
    \includegraphics[width=\textwidth]{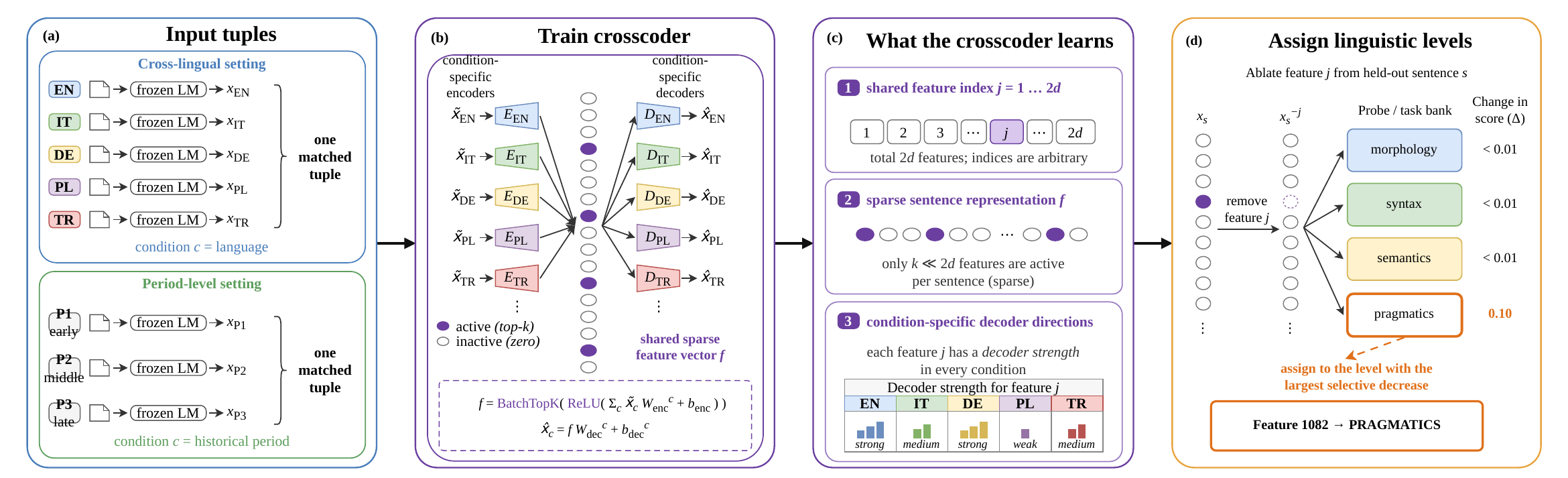}
    \caption{
Overview of crosscoder training and post-hoc linguistic attribution.
\textbf{(a)} Matched cross-lingual or diachronic sentence tuples are constructed by sampling stratum.
\textbf{(b)} Condition-specific encoders and decoders learn shared sparse features through reconstruction alone.
\textbf{(c)} The trained model yields aligned but linguistically unlabeled features and decoder directions.
\textbf{(d)} Held-out feature ablations assign each feature to the linguistic level with the largest selective prediction drop.
}

    \label{fig:crosscoder-main}
\end{figure*}

We introduce \chronolens{}, a framework for studying how multilingual representations change over time. The framework connects three dimensions of variation: language, historical period, and linguistic level. Figure \ref{fig:crosscoder-main} summarizes the \chronolens{} pipeline.
It measures how far each language moves through the learned feature space, 
whether pairs of languages move in similar directions, and whether the four linguistic levels exhibit aligned trajectories within a language.
Each sentence is encoded once with a frozen multilingual language model, and the resulting representation is used throughout the pipeline. We train crosscoders under two comparison conditions: a cross-lingual setting that contrasts languages and a period-level setting that contrasts historical periods within one language. The crosscoders learn sparse features without linguistic supervision; only after training do we assign these features to morphology, syntax, semantics, or pragmatics. This separation prevents the linguistic labels from shaping the learned feature inventory.

\subsection{Representation Learning and Linguistic Attribution}

\noindent\textbf{Input tuples.}
The primary analysis uses parliamentary sentences sampled independently of predefined target words. Cross-lingual comparisons cover 1950--2020, the period available for all five languages, whereas within-language analyses use the full historical record available for each language.
We construct input tuples by matching sentences on token length. The matched sentences are neither translations nor paraphrases. Length matching controls for systematic differences across languages and periods that could otherwise allow the crosscoder to distinguish conditions from sentence length rather than linguistic content.
In the cross-lingual setting, each tuple contains one matched sentence from each language. In the period-level setting, each tuple contains matched sentences from different historical periods of the same language (Fig.~\ref{fig:crosscoder-main}(a)).

\noindent\textbf{Contextual representations.}
Using pretrained multilingual LLMs,
we encode each sentence in an input tuple as a contextualized representation (Fig.~\ref{fig:crosscoder-main}(a)).\footnote{Refer to Appendix~\ref{app:backbones} for more details.
}
Our analysis uses four multilingual backbones:
Qwen3-8B \citep{yang2024qwen3}, Llama-3.1-8B
\citep{dubey2024llama3}, Mistral-Nemo-2407
\citep{jiang2023mistral,MistralAI}, and EuroLLM-9B-2512
\citep{martins2024eurollm}. We select these models for their broad language coverage and their variation in model family, tokenizer, and pretraining data.  Agreement across them is therefore less likely to result from one model's training procedure. EuroLLM provides an additional contrast because it was developed specifically for European languages.

\noindent\textbf{Crosscoder training.}
\label{sec:crosscoders}
Using the contextualized representations from each multilingual backbone, we train a crosscoder that learns a shared feature index together with a separate decoder for each language or period
\citep{lindsey2024crosscoders,
minder2026overcomingsparsityartifactscrosscoders} (Fig.~\ref{fig:crosscoder-main}(b)).
We train each crosscoder on 20k input tuples per decade and language, without linguistic labels, and reserve 10\% for validation. Sampling may reuse sentences when a stratum contains too few distinct instances; we therefore report both the number of tuples and the number of distinct sentences. Appendix \ref{app:concepts} provides the complete sampling procedure.
For an input tuple
$\{\widetilde{\mathbf{x}}_c\}_{c=1}^{C}$,
where $c$ indexes a condition, either a language in the cross-lingual setting or a historical period in the period-level setting, the crosscoder computes

\begingroup
\small
\begin{align}
    \mathbf{f}
    &=
    \operatorname{BatchTopK}\!\left(
    \operatorname{ReLU}\!\left(
        \sum_{c=1}^{C}
        \widetilde{\mathbf{x}}_c
        \mathbf{W}_{\mathrm{enc}}^{c}
        +
        \mathbf{b}_{\mathrm{enc}}
    \right)\right),
    \\
    \widehat{\mathbf{x}}_c
    &=
    \mathbf{f}\mathbf{W}_{\mathrm{dec}}^{c}
    +
    \mathbf{b}_{\mathrm{dec}}^{c}.
\end{align}
\endgroup


The encoder projects each condition-specific representation
into a shared feature space and sums the projected representations across the tuple.
After adding the shared bias, ReLU removes negative activations and BatchTopK retains only the strongest positive activations. The resulting vector $\mathbf{f}$ is thus one sparse representation whose feature indices are shared across all conditions (Fig.~\ref{fig:crosscoder-main}(c)). Each condition-specific decoder maps this vector back into the representation space of condition $c$. 
The shared feature vector aligns feature identity across conditions, whereas the separate decoders allow the same feature to contribute differently to each language or period.
We set the dictionary size, i.e., the number of learned sparse features,
to twice the backbone hidden dimension and use a BatchTopK target active fraction of 0.10. Appendix~\ref{app:crosscoder-details} reports the training objective, sparsity settings, and reconstruction checks.

\noindent\textbf{Linguistic level assignment.}
We assign crosscoder features to linguistic levels using a suite of 23 complementary sentence-level tasks. The five \textbf{morphology} tasks characterize the main predicate through tense, mood, voice, inflectional load, and deverbal nominalization density, drawing on prior work on multilingual morphological complexity, compositional morphology, and grammaticalization \citep{berdicevskis-etal-2018-using,cotterell-schutze-2018-joint,nagata-etal-2026-cross}.
The six \textbf{syntax} tasks measure clause embedding, length-adjusted tree depth, dependency distance, head direction, nominal modification, and coordination, all established properties of dependency structure and diachronic syntactic variation \citep{liu-etal-2022-dependency,krielke-etal-2025-tracing,chen-etal-2026-syntactic}.
To avoid equating sentence meaning with policy topic, the six \textbf{semantic} tasks characterize predicate and argument types, abstractness, negation, quantification and policy-frame task \citep{otmakhova-frermann-2025-narrative}.
The six \textbf{pragmatics} tasks cover deixis, stance, modality, communicative act, evidentiality, and politeness, following work that treats political language use as more than sentence form alone \citep{reinig-etal-2024-politics,subramanian-etal-2019-target}.
Labels are derived primarily from Universal Dependencies parses and multilingual lexical resources \citep{nivre2020ud,qi2020stanza}; continuous measures are discretized into low, medium, and high classes. Semantic and pragmatic labels use deterministic weak-supervision rules that combine parse-derived morphosyntactic and dependency cues with curated multilingual lexicons for predicate and argument types, quantification, modality, stance, evidentiality, and forms of address.
At the morphological level, ``The proposal was rejected'' illustrates passive voice. 
At the syntactic level, ``The bill was debated and amended'' illustrates coordination, while ``the proposed fiscal reforms'' illustrates nominal modification. 
At the semantic level, ``We do not support the proposal'' illustrates negation, and ``Every member supported the motion'' illustrates quantification. 
At the pragmatic level,``I believe we should act'' illustrates personal deixis, while ``The government must respond'' illustrates modality. These examples illustrate the intended linguistic constructs, the actual task labels are derived automatically from dependency parses and multilingual lexical resources.

After crosscoder training, we fit a linear probe for every language-task pair and compare it with an otherwise identical permuted-label control \citep{hewitt-liang-2019-designing}. Because probing is performed only after feature learning, these labels interpret the learned representation without influencing the feature inventory.
For each held-out sentence, we ablate one crosscoder feature at a time and recompute the correct-label probability under every task probe (Fig.~\ref{fig:crosscoder-main}(d)). We aggregate the resulting probability decreases over the tasks belonging to each linguistic level. A feature is assigned to the level with the largest aggregate decrease only when that effect satisfies the attribution threshold and clearly exceeds its effects on the other levels; otherwise, the feature remains unassigned. Because these levels are not mutually exclusive, the
assignment denotes a feature's dominant selective effect rather than an
exclusive linguistic interpretation and prevents double counting across
level-specific trajectories. This procedure provides a post-hoc linguistic attribution of the shared feature dictionary. Appendix~\ref{app:feature-interventions} reports the thresholds and random-direction controls.


\subsection{Evaluation and Diachronic Measurements}

\noindent\textbf{Baselines and measurement validation.}
\label{sec:baselines}
To assess the quality of the learned crosscoder features, we compare them with two controlled baselines. The
\emph{embedding} \cite{hamilton2016diachronic,hagen-2025-lexical} baseline uses the frozen sentence representations directly; for
feature-level evaluation, we use their first 512 principal components. The
\emph{pooled SAE} \cite{andrylie2025sparse,karvonen2025saebench} learns a shared sparse dictionary from representations pooled across all five languages but uses one decoder for each language. In contrast, the crosscoder jointly encodes condition-specific inputs into a shared feature vector and reconstructs them with condition-specific decoders.

We evaluate four properties commonly used to assess sparse representations
\citep{gao2025scaling,balagansky-etal-2025-train, karvonen2025saebench,kantamneni2025sparseprobing}:
\emph{(1) Reconstruction} is the fraction of held-out activation variance left
unexplained by the reconstruction, with lower Fraction of Variance Unexplained (FVU) indicating greater fidelity; it is undefined for uncompressed embeddings. 
\emph{(2) Trajectory stability} is the mean cosine between the full-data displacement vector and vectors obtained by resampling sentences within each language--decade cell, so higher values indicate that the estimated direction does not depend strongly on the sampled sentences. 
(3) \textit{Linguistic agreement} is the mean Spearman correlation between decade-to-decade representational displacement and changes in directly measured linguistic indicators. 
These indicators are computed from the same underlying sentences and partly share the UD annotations and lexical resources used for post-hoc feature attribution.
Linguistic agreement therefore measures correspondence with our operational linguistic indicators, rather than validation against an independent historical gold standard.
\emph{(4) Linguistic specificity} is the proportion of a feature's total
level-aggregated ablation effect concentrated on its most affected linguistic level; $0.25$ corresponds
to equal effects across the four levels, whereas larger values indicate more
level-selective features. We evaluate these effects against permuted-label
probe controls following \citet{hewitt-liang-2019-designing}.


\noindent\textbf{Measuring historical change.}
\label{sec:trajectories}
We compute a separate trajectory for each language and linguistic level. Let
$D_{\ell,t}$ be the set of sentences in language $\ell$ and decade $t$,
$\mathbf{f}_s$ the sparse feature vector of sentence $s$, 
and
$S_\lambda$ the features assigned to linguistic level $\lambda$. The
representation of language $\ell$ at level $\lambda$ in decade $t$ is
the mean activation of those features:
\begin{equation}
    \mathbf{u}_{\lambda}(\ell,t)
    =
    \frac{1}{|D_{\ell,t}|}
    \sum_{s\in D_{\ell,t}}
    \mathbf{f}_s[S_\lambda].
    \label{eq:level-representation}
\end{equation}
Here, $\mathbf{f}_s[S_\lambda]$ denotes the entries of
$\mathbf{f}_s$ belonging to level $\lambda$. We compute all quantities
separately for each multilingual backbone and omit the model index for
readability.
For each language, we measure change relative to its first available decade
$t_0^\ell$:
\begin{equation}
    \operatorname{Magnitude}_{\lambda}(\ell,t)
    =
    \frac{
        \left\|
        \mathbf{u}_{\lambda}(\ell,t)
        -
        \mathbf{u}_{\lambda}(\ell,t_0^\ell)
        \right\|_2
    }{
        \left\|
        \mathbf{u}_{\lambda}(\ell,t_0^\ell)
        \right\|_2
    }.
    \label{eq:change-magnitude}
\end{equation}
The numerator is the Euclidean distance from the language's initial
representation, while the denominator normalizes for differences in feature
scale across linguistic levels. A value of $0$ denotes no change from the
initial decade, and larger values denote greater displacement.
To compare directions, let $t_0^{\ell,\ell'}$ and
$t_1^{\ell,\ell'}$ be the first and last decades available 
for both
languages $\ell$ and $\ell'$. Their displacement vectors are
\begin{equation}
    \boldsymbol{\Delta}_{\lambda}^{\ell}
    =
    \mathbf{u}_{\lambda}(\ell,t_1^{\ell,\ell'})
    -
    \mathbf{u}_{\lambda}(\ell,t_0^{\ell,\ell'}),
\end{equation}
with $\boldsymbol{\Delta}_{\lambda}^{\ell'}$ defined analogously. We measure
directional alignment using cosine similarity:
\begin{equation}
    \operatorname{Direction}_{\lambda}(\ell,\ell')
    =
    \frac{
        \boldsymbol{\Delta}_{\lambda}^{\ell}
        \cdot
        \boldsymbol{\Delta}_{\lambda}^{\ell'}
    }{
        \left\|\boldsymbol{\Delta}_{\lambda}^{\ell}\right\|_2
        \left\|\boldsymbol{\Delta}_{\lambda}^{\ell'}\right\|_2
    }.
    \label{eq:change-direction}
\end{equation}
Values near $1$ indicate parallel change, values near $-1$ indicate
change in opposite directions, and values near $0$ indicate unrelated
directions. Magnitude therefore measures how far each language moves, whereas
direction measures whether two languages move similarly.

\section{Results}
\label{sec:results}

All results use the same matched sentence samples, the same four backbones, and the same
crosscoder configuration described in \S\ref{sec:method}; every reported value is the mean
over the four measuring models, with their spread reported alongside. Magnitudes follow
Eq.~\ref{eq:change-magnitude} 
and directions follow Eq.~\ref{eq:change-direction}.
We first test whether the crosscoder provides a more linguistically valid representation
than dense embeddings and a pooled sparse autoencoder (\S\ref{sec:results-validity}).
We then examine the magnitude and temporal profile of change in each language
(\S\ref{sec:results-magnitude}). Finally, we test whether similar magnitudes imply similar
directions, both across languages and across linguistic levels
(\S\ref{sec:results-direction}).

\subsection{Crosscoders recover more linguistically grounded features}
\label{sec:results-validity}

\begin{table}[!ht]
\centering
\small
\resizebox{\linewidth}{!}{
\begin{tabular}{@{}lccc@{}}
\toprule
\textbf{Criterion}
& \textbf{Embeddings}
& \textbf{SAE}
& \textbf{Crosscoder} \\
\midrule
Held-out reconstruction, FVU $\downarrow$
& \textit{n/a}
& $0.07$
& $\mathbf{0.04}$ \\

Trajectory stability $\uparrow$
& $0.90$
& $0.91$
& $\mathbf{0.92}$ \\

Linguistic agreement, $\rho\uparrow$
& $0.29$
& $0.28$
& $\mathbf{0.72}$ \\

Linguistic specificity, $\uparrow$
& $0.74$
& $0.73$
& $\mathbf{0.89}$ \\
\bottomrule
\end{tabular}}
\caption{
Controlled comparison of dense embeddings, a SAE, and the crosscoder.
FVU is not applicable to embeddings because they do not reconstruct a compressed representation.
}
\label{tab:validity}
\end{table}

Table~\ref{tab:validity} shows that the crosscoder's main advantage is
linguistic rather than geometric. Agreement with direct changes in tense,
subordination, policy-topic, and speech-act distributions increases to
$\rho=0.72$, compared with $0.29$ for embeddings and $0.28$ for the pooled
SAE. 
Linguistic specificity similarly increases to $0.89$, from $0.74$ and
$0.73$. In contrast, the pooled SAE does not improve over the original
embeddings on either measure. A shared sparse dictionary is therefore not
sufficient by itself. The improvement appears when shared feature identities
are combined with condition-specific decoders.
The comparison is less differentiated on reconstruction and stability. The
crosscoder reduces FVU from $0.07$ to $0.04$, but all three representations
produce highly stable directions, with bootstrap cosines between $0.90$ and
$0.92$. Thus, the crosscoder does not obtain its linguistic advantage by
producing substantially smoother trajectories. It preserves the stable signal
already present in the representations while organizing it into features that
better correspond to linguistic variation.
Table~\ref{tab:cc-examples-migration-en} examines four representative features
from the English analysis, showing how the learned inventory distinguishes
historical periods, linguistic levels, and recurring lexical or grammatical
patterns.
\begin{table}[t]
\centering
\scriptsize
\setlength{\tabcolsep}{2.5pt}
\renewcommand{\arraystretch}{1.12}
\begin{tabularx}{\columnwidth}{
    @{}r
    l
    l
    >{\raggedright\arraybackslash}X
    @{}
}
\toprule
\textbf{Feat.} &
\textbf{Level} &
\textbf{Peak} &
\textbf{Top-activating excerpt} \\
\midrule

2373 &
Morph. &
P1 &
\colorbox[rgb]{1.00,0.79,0.67}{He}
\colorbox[rgb]{1.00,0.51,0.25}{thought}
it
\colorbox[rgb]{1.00,0.45,0.15}{unwise}
to leave them \ldots permanently alienated. \\[2pt]

5236 &
Prag. &
P2 &
\colorbox[rgb]{1.00,0.59,0.37}{asked}
the
\colorbox[rgb]{1.00,0.69,0.52}{Minister}
\colorbox[rgb]{1.00,0.45,0.15}{whether}
it is the intention of the Government \ldots ? \\[2pt]

7333 &
Sem. &
P4 &
those whose
\colorbox[rgb]{1.00,0.71,0.55}{asylum}
\colorbox[rgb]{1.00,0.62,0.41}{cases}
are
\colorbox[rgb]{1.00,0.65,0.46}{outstanding}
are
\colorbox[rgb]{1.00,0.59,0.37}{detained}
\ldots \\[2pt]

1082 &
Prag. &
P4 &
\colorbox[rgb]{1.00,0.55,0.30}{Does}
he not
\colorbox[rgb]{1.00,0.45,0.15}{agree}
that
\colorbox[rgb]{1.00,0.87,0.80}{immigration}
creates problems? \\

\bottomrule
\end{tabularx}
\caption{
Representative period-specific features from the English
crosscoder. \textbf{Peak} reports the dominant period: P1 = 1803--1899, P2 = 1900--1945, P3 = 1946--1979,
P4 = 1980--2004, and P5 = 2005--2026. Linguistic levels are assigned by
held-out feature ablation rather than inspection of the examples. Shading
shows activation intensity within each excerpt.
}
\label{tab:cc-examples-migration-en}
\end{table}
Table~\ref{tab:cc-examples-migration-en} shows that period specificity is not
equivalent to topic specificity. Feature 5236 captures the P2 written-question
construction \emph{asked the Minister whether}; 
its strongest activations span
different policy content, indicating a pragmatic parliamentary frame rather
than a migration subtopic. Feature 1082 captures a later pragmatic form,
\emph{Does he not agree that}, associated with adversarial oral questioning
in P4. 



\subsection{Historical change is comparable in magnitude across levels but differs across languages}
\label{sec:results-magnitude}

\begin{figure}[!ht]
    \centering
    \includegraphics[width=\linewidth]{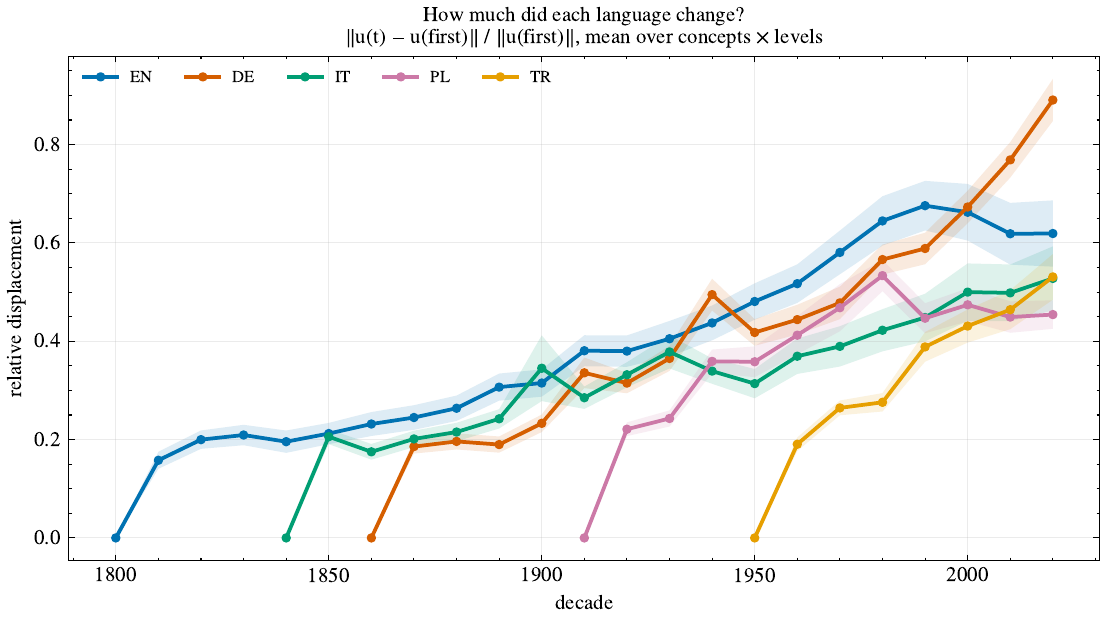}
    \caption{
    Magnitude of historical change over each language's available record.
    At decade $t$, the trajectory reports
    $\operatorname{Magnitude}_{\lambda}(\ell,t)$ relative to the language's
    first available decade, averaged over linguistic levels.
    Shaded intervals show variation across the four measuring models.
    Because the first available decade differs by language, the figure
    compares complete within-language trajectories rather than a common
    historical interval.
    }
    \label{fig:change-over-time}
\end{figure}

\begin{figure}[!ht]
    \centering
    \includegraphics[width=\linewidth]{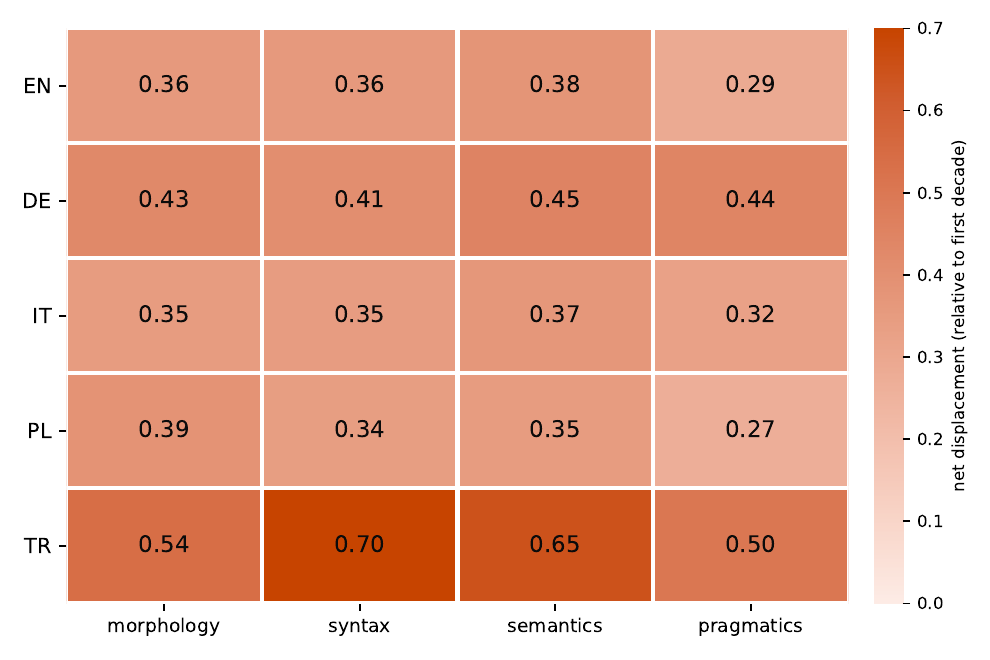}
    \caption{
    Magnitude of historical change between 1950 and 2020 by language and
    linguistic level. All languages are evaluated over the same historical
    interval. Darker cells indicate greater
    magnitude of historical change from the 1950 representation.
    }
    \label{fig:magnitude-by-level}
\end{figure}

Figure~\ref{fig:change-over-time} reports the magnitude of historical change
relative to each language's first available decade. The languages differ in
both their final magnitude and their temporal profile. German has the largest
final magnitude indicating largest change, 
reaching $0.89$ after a marked increase from approximately
1980 onward. English increases more gradually, reaches its highest magnitude
in the 1990s, and ends at $0.62$. Polish reaches approximately $0.53$ around
1980 but decreases to $0.45$ by the final decade. Italian and Turkish both end
at $0.53$, although their trajectories cover different periods and develop
differently over time. Italian and Turkish have the same final value but different temporal
profiles, while the final values for English and Polish are lower than their
earlier maxima. Comparisons across languages must therefore consider both the
magnitude at a given decade and the trajectory through which that magnitude
develops. 


Figure~\ref{fig:magnitude-by-level} compares the magnitude of change over the
common 1950--2020 interval. Averaged across linguistic levels, Turkish has
the highest magnitude at $0.61$, followed by German at $0.43$. English and
Italian both average $0.35$, while Polish averages $0.34$. This ranking
differs from the full-record comparison in
Figure~\ref{fig:change-over-time}: German has the highest magnitude over its
complete record, whereas Turkish has the highest magnitude within the common
1950--2020 period. Magnitude comparisons therefore depend on the historical
interval used.
Variation across languages is larger than variation across linguistic
levels. 
The Turkish syntax and semantics cells also show the largest
between-model spreads. This variation may partly reflect model-specific
tokenization of Turkish: its agglutinative morphology can produce different
subword segmentations across tokenizers, and such differences can affect
morphology-sensitive evaluations
\citep{arnett-bergen-2025-language,basar-bisazza-2026-morphology}. 
An analysis of individual linguistic measures provides a more
direct interpretation of the aggregate magnitudes
(Appendix~\ref{app:observable-change}). We examine 23 measures grouped under
morphology, syntax, semantics, and pragmatics. Across five languages, 37 of the
90 fitted trends remain significant.
Personal deixis increases in all five languages, significantly in German,
Italian, Polish, and Turkish. Passive voice decreases in English, German, and
Turkish but increases in Polish. Long dependencies decrease in English, German,
and Italian but increase in Polish.
ross-language agreement is strongest for pragmatics, with a mean pairwise cosine similarity of $0.92$, while syntax has no common overall direction, with a mean of $0.00$. These results show that similar aggregate magnitudes can result from different changes in the underlying linguistic measures.

\paragraph{Agreement with historical linguistic evidence.}
We further compare our findings with documented historical
linguistic changes. The observed decline in English
\textit{be}-passives, increase in German gender-fair
\textit{-innen} forms, and decreasing use of the Italian
\textit{passato remoto} are consistent with previous
linguistic research \citep{leech2009change,
acke2019sprachwandel,squartini2000simple}.
Moreover, \textsc{ChronoLens} trajectories show strong
correlations with decade-level linguistic measurements
for German coordination ($r=0.88$), Italian personal
deixis ($r=0.95$), English semi-modals ($r=0.73$),
and German normalized dependency distance ($r=0.76$).
Together, these results provide complementary evidence
that \textsc{ChronoLens} captures meaningful diachronic
variation consistent with previously documented
linguistic developments.

\subsection{Magnitude and direction reveal distinct patterns}
\label{sec:results-direction}

\begin{figure}[!htp]
    \centering
    \includegraphics[width=\columnwidth]{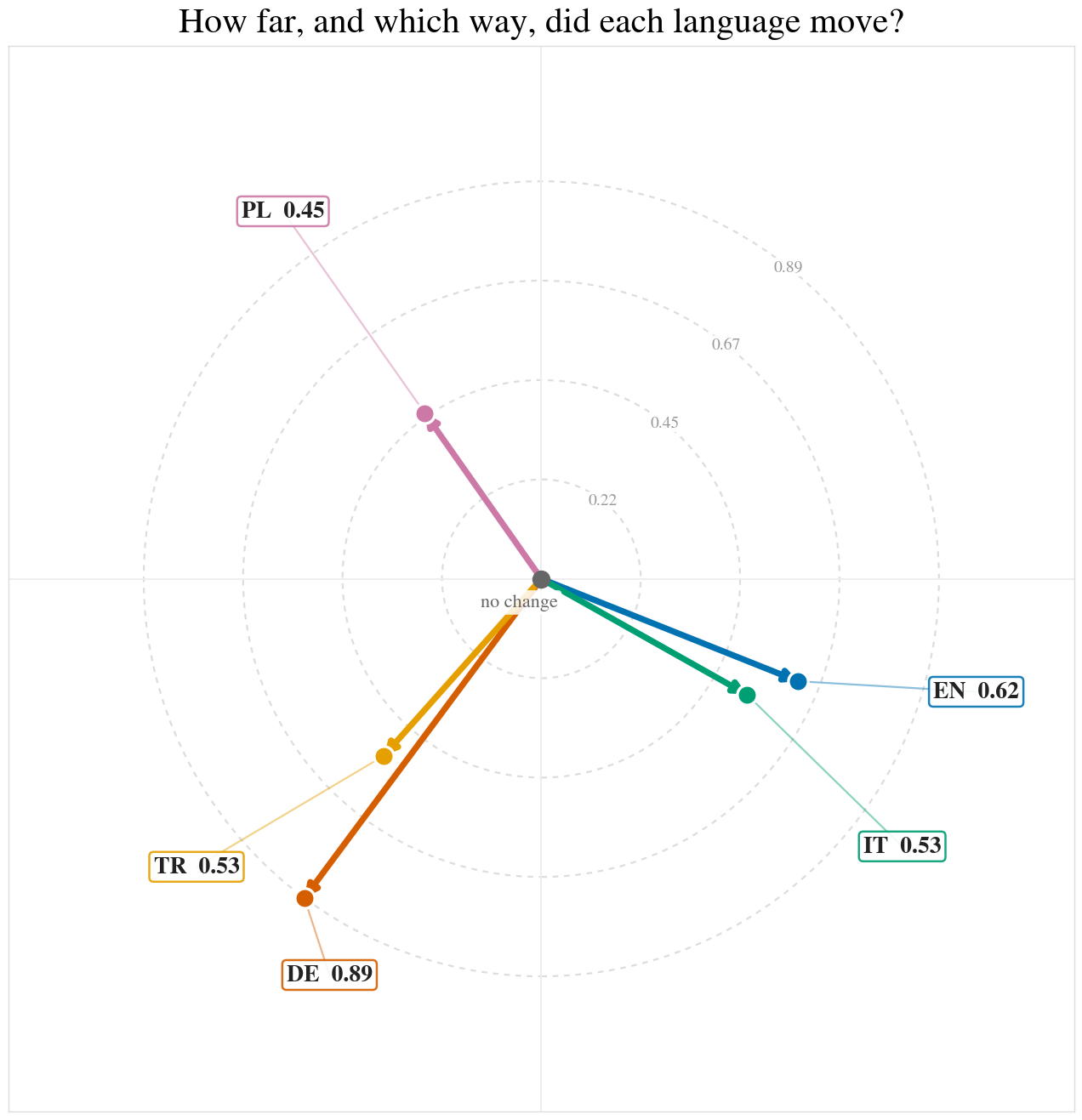}
    \caption{
    Magnitude and direction of historical change over each language's
    available record. The length of each arrow represents the net
    \emph{magnitude of historical change}, computed from the first to the last
    available decade and averaged over linguistic levels. The
    relative angles summarize the \emph{direction} metric, that is, the cosine
    similarity between the corresponding displacement vectors. Languages with
    smaller angular separation have more similar directions of change. 
    }
    \label{fig:change-direction}
\end{figure}

Figure~\ref{fig:change-direction} shows that magnitude and direction capture
different properties of historical change. Italian and Turkish have the same
net magnitude of change, $0.53$, but different directions. German and Turkish
show the opposite pattern: their net magnitudes differ, $0.89$ and $0.53$,
but their directions are similar. The magnitude of change therefore does not
determine the direction of change. 
The directional configuration is not explained by genealogical relatedness
alone. English and Italian have similar directions, as do German and Turkish,
whereas English and German do not form the closest pair despite both being
Germanic languages. This pattern is consistent with the possibility that
parliamentary language responds to shared cultural, political, and
institutional developments. Prior work has distinguished culturally associated
semantic change from language-internal drift and has shown that cultural
differences can be recovered from patterns of language use
\citep{hamilton-etal-2016-cultural,
garimella-etal-2016-identifying}. It also complements studies that identify
language-specific periods of semantic and syntactic change
\citep{hamilton2016diachronic,periti-tahmasebi-2024-systematic,
degaetano-ortlieb-teich-2018-using,krielke-etal-2025-tracing,
chen-etal-2026-syntactic}, as well as work on historical variation in
political framing and parliamentary speech acts
\citep{otmakhova-frermann-2025-narrative,
reinig-etal-2024-politics}. Our results extend these findings by showing that
(i) morphology, syntax, semantics, and pragmatics generally have comparable
magnitudes of change within a language, and (ii) similar magnitudes do not
imply similar directions. Single-level analyses and magnitude alone therefore
capture only part of the observed diachronic structure. 

These measurements support two complementary uses of \chronolens{}. 
First, within-language trajectories identify when historical change is most pronounced and which linguistic levels contribute to it. 
Second, cross-lingual comparisons distinguish the extent of change from its direction. Italian and Turkish exhibit the same overall displacement magnitude (0.53) over their respective historical records, but follow different directions, whereas German and Turkish differ in magnitude while showing similar directions. 
These examples illustrate how \chronolens{} can guide targeted linguistic investigations that would be difficult to motivate using aggregate change magnitudes alone.

\section{Conclusion}
\label{sec:conclusion}
We introduced \chronolens{}, a framework for measuring historical language
change jointly across languages, periods, and linguistic levels. The
crosscoder produces representations that agree more strongly with direct
linguistic stats
while preserving stable historical trajectories.
Across five parliamentary traditions, we find that
morphology, syntax, semantics, and pragmatics generally change by comparable
amounts within a language, but that languages differ substantially in the magnitude, and direction of this change. 
Future work can test whether these patterns generalize
beyond parliamentary discourse and to a broader range of languages,
linguistic measurements, and historical corpora.

\section*{Limitations}

Our division into morphology, syntax, semantics, and pragmatics is also
an analytical simplification. These levels are not mutually exclusive,
and individual phenomena or learned features may span several of them.
The hard assignment used in the trajectory analysis identifies a
feature's dominant probe effect and prevents double counting, but it
can obscure genuinely cross-level features. The resulting trajectories
should therefore be interpreted as changes along four operational
dimensions rather than as a complete decomposition of linguistic
change.
The analysis is restricted to parliamentary discourse in five languages with unequal historical coverage; residual OCR errors, particularly in the earliest material, may still resemble linguistic change despite our filtering. Moreover, frozen multilingual language models remain imperfect measurement instruments whose tokenization, pretraining data, and language coverage may affect the recovered trajectories. 
We evaluate four multilingual backbones independently and aggregate their results, reducing dependence on any single architecture or tokenizer. However, agreement across models does not exclude systematic biases shared by their pretraining distributions. 
In particular, cross-language differences may partly reflect differences in model coverage rather than linguistic change alone \cite{schut2025multilingualllmsthinkenglish}. 
Finally, feature interventions establish relevance to probe predictions, but they do not identify the political, cultural, or institutional causes of the observed changes.

\section*{Broader Impact}

\textsc{ChronoLens} provides a common framework for comparing historical change across languages and linguistic levels, which may support research in computational linguistics, political science, history, and the digital humanities. At the same time, parliamentary records represent institutional discourse produced by political actors rather than the language use of entire populations. Cross-linguistic similarities should therefore not be interpreted as essential properties of national communities or as direct evidence that languages are becoming uniformly more alike. The framework is best used to generate hypotheses that are subsequently evaluated against original texts, dated historical events, and social or institutional evidence. Extending the analysis to additional genres, regions, and less-resourced languages will be important for preventing conclusions about historical language change from being dominated by well-documented European parliamentary traditions.

\section*{Ethical Considerations}

Our analysis uses publicly available parliamentary and political texts from official and research sources and reports aggregate language--period patterns rather than predictions about individual speakers. Nevertheless, parliamentary records may contain identifiable speakers and discussions of sensitive political or social issues. Any release of derived data should preserve source attribution, licensing conditions, and applicable restrictions rather than redistributing source material indiscriminately. The automatic parsers, multilingual lexicons, probes, and pretrained language models used in the pipeline may also encode cultural and language-specific biases; their outputs should be treated as operational measurements, not objective labels or diagnoses of linguistic communities. The resulting representations should not be used for individual political profiling, targeted persuasion, or ranking languages and populations. Reproducibility materials should document data provenance, filtering, sampling, model versions, and known measurement limitations. Finally, we only used LLMs for code generation.

\section*{Acknowledgements}

We thank the anonymous reviewers for their thoughtful
feedback, which helped improve the paper. The NLLG Lab
gratefully acknowledges support from the German Research
Foundation (DFG) through the Heisenberg Grant EG 375/5-1.
This work was supported by the DFG under grant
EG 375/12-1, ``HistoBias: Analyzing Historical and Cross-Cultural
Bias through the lens of NLP''
(\url{https://nl2g.github.io/histobias/}).

\bibliography{custom}

\appendix

\section{Dataset Sources}
\label{app:sources}

\begin{figure}[t]
\centering
\includegraphics[width=\columnwidth]{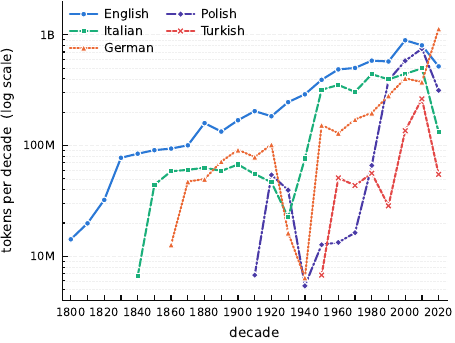}
\caption{Tokens per decade (Calculated using Llama-3 tokenizer, log scale) for each language, 1800s--2020s; colour key
at top left. }
\label{fig:coverage}
\end{figure}

\begin{table*}[t]
\centering
\footnotesize
\resizebox{\linewidth}{!}{
\begin{tabular}{llcrlp{0.28\linewidth}}
\toprule
\textbf{Lang.} & \textbf{Source} & \textbf{Country} & \textbf{Documents} & \textbf{Coverage} & \textbf{Description} \\
\midrule
en & hansard\_historic & GB & 10{,}379{,}346 & 1803--2004 & UK Commons and Lords, Historic Hansard. \\
en & theyworkforyou & GB & 7{,}149{,}348 & 1945--2026 & UK Commons and Westminster Hall debates, TheyWorkForYou. \\
en & parlamint\_gb & GB & 673{,}121 & 2015--2022 & UK component of ParlaMint 5.0. \\
en & manifesto\_uk & GB & 84 & 1964--2024 & UK party manifestos from the Manifesto Project. \\
\midrule
pl & ppc\_pl & PL & 13{,}898{,}425 & 1919--2025 & Polish Parliamentary Corpus, Sejm and Senate. \\
pl & parlamint\_pl & PL & 229{,}012 & 2015--2022 & Polish component of ParlaMint 5.0. \\
pl & sejm\_pl & PL & 148 & 2015--2023 & Sejm proceedings and interpellations. \\
pl & manifesto\_pl & PL & 43 & 1991--2019 & Polish party manifestos from the Manifesto Project. \\
\midrule
it & itaparl & IT & 5{,}639{,}906 & 1948--2022 & Camera dei Deputati speech turns, ItaParl. \\
it & parlamint\_it & IT & 174{,}184 & 2013--2022 & Italian component of ParlaMint 5.0. \\
it & italian\_parliament & IT & 34{,}857 & 1848--2022 & Camera and Senato, Kingdom of Italy through Republic, full-text OCR. \\
it & manifesto\_it & IT & 112 & 1963--2018 & Italian party manifestos from the Manifesto Project. \\
\midrule
de & parlamint\_de\_beta & DE & 2{,}684{,}792 & 1949--2025 & ParlaMint-DE beta, Bundestag debates. \\
de & reichstag\_bsb & DE & 2{,}645{,}175 & 1867--1939 & Reichstagsprotokolle, full-text OCR. \\
de & germaparl & DE & 1{,}042{,}888 & 1949--2021 & GermaParlTEI Bundestag debates. \\
de & bundestag\_official & DE & 620{,}975 & 2025--2026 & Bundestag DIP API, official recent records. \\
de & parlamint\_at & AT & 232{,}980 & 1996--2022 & Austrian parliament, German-language ParlaMint component. \\
de & ddb\_newspapers & DE & 100{,}000 & 1945--2024 & German Digital Library historical newspapers. \\
de & manifesto\_de & DE & 104 & 1949--2025 & German party manifestos from the Manifesto Project. \\
\midrule
tr & tbmm\_speeches\_v1 & TR & 1{,}207{,}674 & 1950--2023 & Turkish Grand National Assembly linked corpus. \\
tr & parlamint\_tr & TR & 682{,}387 & 2011--2022 & Turkish component of ParlaMint 5.0. \\
tr & manifesto\_tr & TR & 28 & 1954--2018 & Turkish party manifestos from the Manifesto Project. \\
\bottomrule
\end{tabular}}
\caption{All sources in the corpus. Counts are post-deduplication counts in the unified record schema. The main analyses use parliamentary speech; manifestos and German newspapers are retained for corpus breadth.}
\label{tab:sources-full}
\end{table*}

\section{Dataset examples}
\label{app:example}
Every one of the 22 sources in Table~\ref{tab:sources-full} is normalised onto the single
16-field record schema described in \S\ref{sec:dataset}. Table~\ref{tab:example-record} shows one
complete record and
Table~\ref{tab:example-multiling} shows one record per language, drawn from each language's
principal parliamentary source. Together they make concrete what the cross-lingual comparison
rests on: the
\texttt{lang}, \texttt{year} and \texttt{doc\_type} fields carry the same meaning in every
collection, so a metric computed over English speech is computed over the same kind of object as
the same metric over Italian or Turkish speech.

\begin{table}[!h]
\centering\small
\resizebox{\columnwidth}{!}{
\begin{tabular}{@{}p{0.26\columnwidth}p{0.68\columnwidth}@{}}
\toprule
\textbf{Field} & \textbf{Value} \\
\midrule
\texttt{id} & \texttt{germaparl\_a98144794f94707d} \\
\texttt{source} & \texttt{germaparl} \\
\texttt{country} & \texttt{DE} \\
\texttt{lang} & \texttt{de} \\
\texttt{doc\_type} & \texttt{parliamentary\_speech} \\
\texttt{date} & \texttt{1994-05-26} \\
\texttt{year} & \texttt{1994} \\
\texttt{title} & \texttt{Bundestag 1994-05-26 WP12 Nr230} \\
\texttt{speaker} & \texttt{Gerd Wartenberg (Berlin)} \\
\texttt{speaker\_role} & \texttt{NA} \\
\texttt{party} & \texttt{SPD} \\
\texttt{house} & \texttt{Bundestag} \\
\texttt{granularity} & \texttt{speech} \\
\texttt{url} & \textit{(null)} \\
\texttt{meta} & \texttt{\{"leg": "12", "sess": "230"\}} \\
\texttt{text} & Herr Staatssekret\"{a}r Lintner, meinen Sie denn --- bei aller Kritik an diesem Gesetz ---, da\ss{} es sinnvoll ist, diese alten H\"{u}te wieder hervorzuholen und diesen Streit fortzusetzen: Sind wir ein Einwanderungsland oder kein Einwanderungsland? Glauben Sie, da\ss{} uns das weiterf\"{u}hrt? Es sind in die Bundesrepublik Deutschland 6 Millionen Menschen zugewandert. K\"{o}nnen wir nicht einmal von dieser Basis ausgehen, diese Menschen hier vern\"{u}nftig zu integrieren [\ldots{}] (Beifall bei der SPD und dem B\"{U}NDNIS 90/DIE GR\"{U}NEN) \\
\bottomrule
\end{tabular}}
\caption{One complete record from the unified corpus (source \emph{germaparl}, German
Bundestag, 26 May 1994), abridged only in \texttt{text}.\texttt{meta} holds the source-specific fields that do not fit the shared
schema and is stored as a JSON string so the Parquet schema stays byte-identical across sources;
here it records the legislative period and sitting number. Fields absent at source are null
(\texttt{url}) or carry the source's own missing-value token (\texttt{speaker\_role} =
\texttt{NA}).}
\label{tab:example-record}
\end{table}

\begin{table*}[!h]
\centering\footnotesize
\setlength{\tabcolsep}{4pt}
\begin{tabular}{@{}llcp{0.19\textwidth}p{0.47\textwidth}@{}}
\toprule
\textbf{Lang.} & \textbf{Source} & \textbf{Year} & \textbf{Speaker / Party} & \textbf{\texttt{text} (excerpt)} \\
\midrule
en & \emph{hansard\_historic} & 1803 & The Speaker / --- & acquainted the House that the House h3d, in obedience to his Majesty's command, attended in the House of Peers, to hear his Majesty's most gracious Speech from the Throne [\ldots{}] \\
de & \emph{germaparl} & 1994 & G. Wartenberg / SPD & Sind wir ein Einwanderungsland oder kein Einwanderungsland? [\ldots{}] Es sind in die Bundesrepublik Deutschland 6 Millionen Menschen zugewandert. [\ldots{}] \\
it & \emph{itaparl} & 1950 & O. L. Scalfaro / DC & SCALFARO. Voi non credete nella Costituzione; ne usate come tappa per la conquista violenta del potere. (Interruzione del deputato Grilli). [\ldots{}] Dove vie liberta nel giudizio, si ha affermazione di giustizia [\ldots{}] \\
pl & \emph{ppc\_pl} & 1919 & F. Radziwi\l{}\l{} / KPK & Wysoka Izbo! W chwili gdy przypada na mnie obowi\k{a}zek zagajenia tego Wysokiego Zgromadzenia, opr\'{o}cz serdecznej rado\'{s}ci czuj\k{e} wzruszenie [\ldots{}] \\
tr & \emph{tbmm\_speeches\_v1} & 2018 & M. A. Kaya / AKP & Say{\i}n Ba\c{s}kan{\i}m, tabii ki konu\c{s}mac{\i}n{\i}n \"{o}zellikle ortaya koymu\c{s} oldu\u{g}u ger\c{c}ek d{\i}\c{s}{\i} beyanlar{\i}na girmeden \"{o}nce \c{s}undan bahsetmek isterim [\ldots{}] \\
\bottomrule
\end{tabular}
\caption{One record per language, each taken from that language's principal parliamentary
source, showing the same 16-field schema realised across all five.}
\label{tab:example-multiling}
\end{table*}

\paragraph{Data schema.} Three properties of the release are worth stating
explicitly, because they constrain how the records may be used. First, \texttt{granularity}
distinguishes a \emph{speech} (one uninterrupted turn) from a \emph{segment} (a chunk of a longer
document, suffixed \texttt{\#seg}$k$ in the \texttt{id}) and from a \emph{session} (a whole
sitting); metrics that assume a single speaker must be restricted to \texttt{granularity} =
\texttt{speech}. Second, \texttt{date} is a string and may be partial---\texttt{1919} for an OCR'd
annual volume, \texttt{1964-10} for a manifesto, a full \texttt{YYYY-MM-DD} for a modern sitting---
so the integer \texttt{year} field, not \texttt{date}, is the safe key for the decade binning used
throughout. Third, \texttt{id} is inherited from the originating source and is a document
identifier, not a primary key: in the session-structured sources (\emph{germaparl},
\emph{bundestag\_official}, \emph{theyworkforyou}) every speech of one sitting shares its sitting's
identifier, so records must be addressed by row rather than looked up by \texttt{id}.

\section{\chronolens{} methodological details}
\label{app:methods}

\subsection{Concept matching and sentence sampling}
\label{app:concepts}

\paragraph{Concept inventory.}

The analysis covers 13 concepts. Migration and gender are the target
concepts. Defence, democracy, economy, environment, Europe, religion,
security, taxation, and technology represent general political discourse.
Road and water are frequency-matched controls. We keep the three groups
separate because a pattern found for all political concepts may reflect a
change in parliamentary discourse generally, while a pattern also found for
road and water is more likely to reflect corpus or model variation.

Each concept is represented by a language-specific set of lowercased stems.
A stem is matched at the beginning of a word, with longer alternatives tested
first. Prefix matching is used because exact word matching would miss common
inflected forms, particularly in German, Polish, and Turkish. It may also
include words that share a stem but not the intended sense. We therefore
report the full lexicons with the released data and inspect ambiguous terms
during error analysis.

A sentence is assigned to at most one concept. When several concepts occur
in the same sentence, the first matched concept determines the assignment.
This rule prevents the same sentence from contributing to several concept
trajectories, although it means that concept counts are a partition of the
harvested sentences rather than an estimate of their total corpus frequency.

\paragraph{Concept masking.}

For every matched sentence, we retain the original text and create a masked
version in which the concept expression and its inflectional ending are
replaced with \texttt{<concept>}. The policy-frame labels and matching strata
are calculated from the masked text. During activation extraction, we also
exclude the concept span from mean pooling. These two operations prevent the
concept term itself from determining either the matching stratum or the model
representation.

\paragraph{Matched strata.}

We divide sentences into strata based on token length and dominant policy
frame. Length is divided into four bands: fewer than 12 tokens, 12 to 24
tokens, 25 to 44 tokens, and at least 45 tokens. The policy frame is selected
from a fixed multilingual inventory that covers domains such as economy,
migration, security, rights, welfare, employment, environment, law,
education, religion, and foreign policy.

Cross-lingual and temporal tuples are sampled only within a shared stratum.
The sentences in a tuple therefore have similar lengths and broad policy
content. This matching does not make them translations or paraphrases, but it
removes two simple sources of variation that the crosscoder could otherwise
use to distinguish conditions.

\paragraph{Temporal balancing.}

We retain at most 400 sentences for each concept, language, and decade. The
cap prevents recent decades from dominating the analysis because they contain
more digitized material. Reservoir sampling with a fixed seed is used so that
the selected sentences remain reproducible.
A decade is used as a crosscoder condition only when it contains at least 250
distinct sentences after matching. The final data provide approximately 293
distinct sentences per decade condition, 855 per period-level condition, and
3,187 per cross-lingual condition. These values refer to distinct sentence
representations, not the larger number of tuples obtained by resampling them.

\subsection{Backbones, representations, and layer selection}
\label{app:backbones}

We represent each sentence by mean-pooling the content-token residual stream
from one block of a frozen language model. Extraction uses a single forward
pass under \texttt{no\_grad}, and vectors are stored in fp16. When a target
concept is present, its tokens are excluded from pooling so that changes in its
surface form are not mistaken for contextual change; topic-agnostic sentences
are pooled in full. We use four multilingual backbones, Qwen3-8B,
Llama-3.1-8B, Mistral-Nemo-2407, and EuroLLM-9B-2512, chosen to vary model
family, tokenizer, and pretraining data.

For each backbone, we evaluate layers at approximately 25\%, 50\%, and 75\%
of model depth and select the layer with the highest mean probe selectivity
across four linguistic levels and five languages. Selectivity is defined as
$\mathrm{acc}_{\text{held-out}}(\text{real}) -
\mathrm{acc}_{\text{held-out}}(\text{control})$, where the control probe uses
permuted labels; sentence-length and OCR-noise probes are reported but excluded
from selection. Table~\ref{tab:backbone-layers} gives the selected layers:
L9 for Qwen3-8B, L16 for Llama-3.1-8B, L20 for Mistral-Nemo-2407, and L10 for
EuroLLM-9B-2512. Differences between the best and second-best depths are small
(0.002--0.026), indicating that the linguistic properties are similarly
decodable across the middle layers. We therefore fix one selected layer per
backbone before computing any historical results.

\begin{table*}[t]
\centering
\small
\resizebox{\linewidth}{!}{
\begin{tabular}{@{}lrrccc@{}}
\toprule
Backbone & Blocks & $d_{\text{model}}$ & 25\% & 50\% & 75\% \\
\midrule
Qwen3-8B \citep{yang2024qwen3}
& 36 & 4096
& \textbf{L9: 0.254}$^{\dagger}$
& L18: 0.235
& L27: 0.248 \\

Llama-3.1-8B \citep{dubey2024llama3}
& 32 & 4096
& L8: 0.278
& \textbf{L16: 0.280}$^{\dagger}$
& L24: 0.273 \\

Mistral-Nemo-2407 \citep{jiang2023mistral,MistralAI}
& 40 & 5120
& L10: 0.277
& \textbf{L20: 0.287}$^{\dagger}$
& L30: 0.278 \\

EuroLLM-9B-2512 \citep{martins2024eurollm}
& 42 & 4096
& \textbf{L10: 0.295}$^{\dagger}$
& L21: 0.266
& L32: 0.269 \\
\bottomrule
\end{tabular}}
\caption{Backbone architectures and layer-selection results. The final three columns report mean probe selectivity at approximately 25\%, 50\%, and 75\% of model depth, averaged over the four linguistic levels and five languages. The selected layer for each backbone is shown in bold and marked with $^{\dagger}$.}
\label{tab:backbone-layers}
\end{table*}

\subsection{Crosscoder training and calibration}
\label{app:crosscoder-details}

\paragraph{Input tuples.}

A crosscoder receives matched tuples with one representation from every
condition. In the cross-lingual setting, a tuple contains one representation
from each of the five languages. In the period-level setting, it contains one
representation from each available historical period. In the decade-level
setting, it contains one representation from every decade that meets the
minimum data requirement.

Each crosscoder is trained on 20,000 sampled tuples. We reserve 10\% for
validation, leaving 18,000 training tuples and 2,000 validation tuples.
Sampling may reuse a sentence when the shared strata do not contain enough
distinct examples. We therefore record both the number of sampled tuples and
the number of distinct sentences in every condition.

\paragraph{Condition standardization.}

Activation magnitudes can differ between languages and historical periods.
Without normalization, a condition with larger vector norms could contribute
more to the reconstruction loss. We standardize each condition separately:

\begin{equation}
    \widetilde{\mathbf{x}}_c
    =
    \frac{\mathbf{x}_c-\boldsymbol{\mu}_c}{\sigma_c},
    \qquad
    \sigma_c =
    \sqrt{
        \frac{
            \mathbb{E}
            \left\|
                \mathbf{x}_c-\boldsymbol{\mu}_c
            \right\|_2^2
        }{d}
    },
\end{equation}

where \(c\) denotes a condition and \(d\) is the model hidden dimension. This
normalization preserves directional information while reducing scale
differences between conditions.

\paragraph{Crosscoder objective.}

Following prior work on crosscoders
\citep{jiralerspong2026crossarchitecturemodeldiffingcrosscoders,
minder2026overcomingsparsityartifactscrosscoders}, one sparse feature vector
is inferred jointly from all conditions:

\begin{align}
    \mathbf{f}
    &=
    \operatorname{ReLU}
    \left(
        \sum_{c=1}^{C}
        \widetilde{\mathbf{x}}_c
        \mathbf{W}^{c}_{\mathrm{enc}}
        +
        \mathbf{b}_{\mathrm{enc}}
    \right), \\
    \widehat{\mathbf{x}}_c
    &=
    \mathbf{f}\mathbf{W}^{c}_{\mathrm{dec}}
    +
    \mathbf{b}^{c}_{\mathrm{dec}}.
\end{align}

The feature index is shared across conditions, while each condition has its
own encoder and decoder weights. This structure allows feature \(j\) to be
compared directly across languages or periods.

The dictionary size is set relative to the hidden dimension of the backbone
rather than fixed across models. We use an expansion factor of two, so a
model with hidden dimension \(d\) receives \(2d\) features. This keeps the
degree of overcompleteness comparable across backbones.

\paragraph{Sparsity.}

We use BatchTopK rather than an \(L_1\) penalty. BatchTopK retains the
strongest feature activations in each batch but does not directly penalize
their magnitude. This matters because the shared and condition-specific
classification depends on decoder magnitudes. An \(L_1\) penalty can shrink a
feature unevenly across conditions and make a shared feature appear specific
\citep{minder2026overcomingsparsityartifactscrosscoders}.

The target active fraction is 0.10. An auxiliary reconstruction term revives
features that have remained inactive for 200 optimization steps. During
inference, a threshold estimated from the surviving training activations
removes small activations. Thresholds are calibrated separately for each
condition because a threshold estimated from the joint tuple representation
does not transfer directly to a single condition.

\paragraph{Shared and condition-specific features.}

For feature \(j\), let

\begin{equation}
    n_{c,j}
    =
    \left\|
        \mathbf{W}^{c}_{\mathrm{dec}}[j]
    \right\|_2
\end{equation}

be its decoder norm in condition \(c\). A feature is considered shared when
the ratio between its largest and smallest decoder norms is at most four. It
is considered condition-specific when its largest norm is at least four
times its second-largest norm. Features that meet neither rule are marked as
mixed. Features with negligible total decoder norm are marked inactive.

The main threshold is four, but we also calculate the split at ratios of two,
three, six, and eight. This sensitivity analysis shows whether the conclusion
depends on one boundary.

\paragraph{Latent-scaling check.}

Decoder norms can overstate specificity when training suppresses a shared
direction in one condition or divides one shared mechanism across several
features. For every initially specific feature, we test whether its decoder
direction also explains reconstruction or residual structure in the other
conditions. If it does, the feature is relabeled as shared. This check follows
the latent-scaling analysis of
\citet{minder2026overcomingsparsityartifactscrosscoders}.

\paragraph{Reconstruction value.}

We evaluate reconstruction on the held-out tuples using fraction of variance
unexplained:

\begin{equation}
    \operatorname{FVU}
    =
    \frac{
        \sum_i
        \left\|
            \widetilde{\mathbf{x}}_i
            -
            \widehat{\mathbf{x}}_i
        \right\|_2^2
    }{
        \sum_i
        \left\|
            \widetilde{\mathbf{x}}_i
            -
            \overline{\widetilde{\mathbf{x}}}
        \right\|_2^2
    }.
\end{equation}

An FVU of zero indicates perfect reconstruction, while an FVU of one is no
better than predicting the condition mean. We mark a checkpoint as unusable
when held-out FVU exceeds 0.50 or when any condition falls below the minimum
number of distinct sentences. Feature-level conclusions are not drawn from
unusable checkpoints.

\paragraph{Condition-shuffled null.}

A decomposition may produce apparently specific features even when its
conditions contain no systematic difference. We therefore train a null
crosscoder with the same data, strata, condition count, dictionary size, and
optimization procedure. In the null tuples, every position is sampled from
the pooled condition data, making condition identities exchangeable.

We compare the real and null crosscoders using their number of specific
features and their decoder-norm ratios. The real crosscoder is considered to
resolve its conditions only when it contains at least twice as many specific
features as the null and also meets the minimum feature-coverage requirement.
When it does not pass this test, analyses that depend on the shared-specific
split are marked unsupported rather than interpreted as evidence that the
conditions are identical.

\subsection{Feature attribution and interventions}
\label{app:feature-interventions}



\paragraph{Candidate features.}

For each crosscoder, we score the 1,000 features with the highest activation
mass. Restricting the intervention analysis to these features keeps the
computation manageable and avoids testing features that are rarely active.
The main trajectory analysis is not restricted to features that pass every
diagnostic because the resulting feature set is too small to define stable
trajectory spaces. Instead, the diagnostics are reported as evidence about
the reliability of individual features.

\paragraph{Diagnostic gate.}

The six diagnostics are temporal selectivity, linguistic specificity,
necessity, sufficiency, split-half robustness, and cross-lingual recurrence
where applicable. Their conjunction is stored as a diagnostic indicator. It
is not used to remove features from the main analysis. This choice avoids
constructing trajectories from a very small and uneven set of surviving
features, while still showing how often a feature receives support beyond
correlation.

\paragraph{Linguistic attribution by ablation.}

Let \(f_{s,j}\) be the activation of feature \(j\) on sentence \(s\), and let
\(\boldsymbol{\delta}_{c,j}\) be its decoder direction mapped back to the
original activation scale. We remove the feature only from sentences on which
it is active:

\begin{equation}
    \mathbf{x}^{-j}_s
    =
    \mathbf{x}_s
    -
    f_{s,j}\boldsymbol{\delta}_{c,j}.
\end{equation}

For each linguistic level \(\lambda\), we measure the average decrease in the
probability assigned to the correct label:

\begin{equation}
    D_{\lambda}(j)
    =
    \frac{1}{|A_j|}
    \sum_{s\in A_j}
    \left[
        p_{\lambda}(y_s\mid\mathbf{x}_s)
        -
        p_{\lambda}(y_s\mid\mathbf{x}^{-j}_s)
    \right],
\end{equation}

where \(A_j\) is the set of sentences on which the feature is active. The
feature is assigned to the level with the largest decrease. Measuring only
active sentences prevents a localized feature effect from being diluted by
sentences on which the feature contributes nothing.

We use two criteria to distinguish a selective level attribution from
a small or ambiguous probe effect. First, the largest aggregate
probability decrease must exceed an absolute threshold of 0.01, which
excludes effects that are negligible on the probability scale. Second,
it must be at least 1.2 times the second-largest decrease, which requires
a minimum separation between the two most affected levels. These
criteria implement a deliberately conservative assignment rule:
features with weak effects or comparable effects across several levels
remain unassigned rather than being forced into a single category.
Features active on fewer than 20 held-out sentences also remain
unassigned because their intervention effects cannot be estimated
reliably. For cross-lingual shared features, the final level is
determined by the majority assignment across languages.

\paragraph{Temporal selectivity.}

A feature must first show some relation to historical time. We compare its
activation on early and late sentences using a held-out Mann-Whitney statistic
converted to a direction-independent area under the curve. The feature passes
the temporal-selectivity diagnostic when its AUC is at least 0.55. This
criterion establishes association with time but does not by itself establish
a functional role.

\paragraph{Necessity.}

Necessity asks whether removing a feature weakens the later-period prediction
of the probe associated with its assigned level. We compute the mean decrease
in the relevant probability on later sentences where the feature is active.
The effect must exceed the 95th percentile obtained from random directions
with the same norm.

\paragraph{Sufficiency.}

Sufficiency asks whether adding the feature to earlier representations moves
the probe toward the later-period prediction. We insert the decoder direction
using the feature's mean activation on later firing sentences. As with
necessity, the effect must exceed the matched random-direction null. The
random comparison is used because a large edit to an activation may change a
probe even when the edited direction has no special linguistic role.

\paragraph{Split-half robustness.}

We divide sufficiently large matching strata into two random halves and
calculate the feature's mean activation in each half. A feature passes the
robustness diagnostic when the two activation profiles have a Spearman
correlation of at least 0.4. This test checks whether the feature depends on a
small number of sentences within a stratum.

\paragraph{Cross-lingual recurrence.}

For cross-lingual shared features, we test whether inserting the feature has
the same directional effect in a majority of the other languages. This test
uses the semantics probe because its policy-frame classes have the same
interpretation across the five languages. Tense and subordinate-clause labels
are not fully equivalent across languages, so we do not use them for this
cross-lingual sign test.

\paragraph{Threshold sensitivity.}
We repeat the attribution and trajectory analyses using absolute
thresholds in $\{0.005, 0.01, 0.02\}$ and relative-separation thresholds
in $\{1.1, 1.2, 1.5\}$.
Across these settings the composition of the assigned inventory is
essentially fixed (morphology 11.3--12.3\%, syntax 16.3--17.5\%, semantics 69.7--71.9\%, pragmatics 0.4--0.6\%),
the same 30 of 56 (concept, level) cells clear the minimum-feature criterion, and the
per-cell endpoint convergence indices stay tightly coupled to the main
setting (Pearson $r \geq 0.86$, Spearman $\rho \geq 0.84$, sign agreement 87--100\%).
Pragmatics, however, never reaches that count at any setting (0.4--0.6\% of assigned features), so the filtered trajectories cover morphology, syntax and semantics only.
The pre-registered target~$\times$~decade convergence contrast keeps
its sign and its significance at every setting ($\hat\beta$ = $-0.29$ to $-0.12$, $p \leq 0.020$),
target-concept morphology ($-0.18$ to $-0.02$) and semantics ($-0.10$ to $-0.07$) diverge in all 9 runs (slope per century) while control and political cells converge in 54 of 54,
and the pre-registered verdict is C throughout,
the only sign that moves being target syntax (positive in 6 of 9),
although stricter thresholds reduce the number of assigned features
from 3,581 (at 0.005, 1.1) to 1,297 (at 0.02, 1.5).
We therefore use 0.01 and 1.2 in the main analysis as an intermediate setting
that excludes weak and ambiguous effects while retaining sufficient
feature coverage for stable trajectory estimation.

\begin{table}[t]
\centering
\small
\setlength{\tabcolsep}{4pt}
\begin{tabular}{rrrrrrrl}
\toprule
$\tau_{\mathrm{abs}}$ & $\tau_{\mathrm{rel}}$ & $N$ & cells & $r$ & sign & $\hat\beta$ & $p$ \\
\midrule
0.005 & 1.1 & 3,581 & 30 & 0.90 & 97\% & $-0.12$ & $0.020$ \\
0.005 & 1.2 & 3,507 & 30 & 0.96 & 97\% & $-0.14$ & $0.009$ \\
0.005 & 1.5 & 3,275 & 30 & 0.97 & 93\% & $-0.15$ & $0.006$ \\
0.01 & 1.1 & 2,429 & 30 & 0.93 & 100\% & $-0.18$ & $<10^{-3}$ \\
\textbf{0.01} & \textbf{1.2} & \textbf{2,373} & \textbf{30} & \textbf{1.00} & \textbf{100\%} & $\mathbf{-0.18}$ & $\mathbf{<10^{-3}}$ \\
0.01 & 1.5 & 2,198 & 30 & 0.99 & 100\% & $-0.18$ & $<10^{-3}$ \\
0.02 & 1.1 & 1,436 & 30 & 0.90 & 87\% & $-0.29$ & $<10^{-4}$ \\
0.02 & 1.2 & 1,405 & 30 & 0.91 & 87\% & $-0.25$ & $<10^{-3}$ \\
0.02 & 1.5 & 1,297 & 30 & 0.86 & 87\% & $-0.24$ & $<10^{-3}$ \\
\bottomrule
\end{tabular}
\caption{Sensitivity of level attribution and of the trajectory analysis to criterion~2's absolute floor $\tau_{\mathrm{abs}}$ and relative separation $\tau_{\mathrm{rel}}$. $N$: assigned features (all concepts); cells: (concept, level) cells clearing the minimum feature count; $r$ and sign: Pearson correlation and sign agreement of the per-cell endpoint convergence index against the main setting (bold); $\hat\beta$: target~$\times$~decade convergence contrast (negative $=$ targets diverge relative to controls).}
\label{tab:threshold-sensitivity}
\end{table}

\subsection{Probe diagnostics}
\label{app:probe-diagnostics}

The four probes differ in their number of classes and in the
distribution of their automatically derived labels. Raw accuracy is
therefore not directly comparable across tasks: in particular, a probe
can achieve high accuracy when one class dominates the held-out data.
We report three diagnostics separately for each language--task pair.
\emph{Majority} is the proportion of held-out examples assigned to the
most frequent class; \emph{accuracy} is the probe's held-out accuracy;
and \emph{selectivity} is the difference between held-out accuracy and
the accuracy of the same probe trained on permuted labels
\citep{hewitt-liang-2019-designing}.

\begin{table}[!h]
\centering
\small
\resizebox{\columnwidth}{!}{
\begin{tabular}{@{}llcccc@{}}
\toprule
\textbf{Language}
& \textbf{Task}
& \textbf{Classes}
& \textbf{Majority}
& \textbf{Accuracy}
& \textbf{Selectivity} \\
\midrule
English
& Tense
& 3
& 0.51
& $0.93 \pm 0.01$
& $0.50 \pm 0.00$ \\
English
& Subordination
& 2
& 0.79
& $0.93 \pm 0.01$
& $0.19 \pm 0.00$ \\
English
& Policy topic
& 13
& 0.08
& $0.75 \pm 0.04$
& $0.67 \pm 0.04$ \\
English
& Speech act
& 3
& 0.84
& $0.97 \pm 0.01$
& $0.21 \pm 0.01$ \\
\midrule
German
& Tense
& 3
& 0.65
& $0.87 \pm 0.01$
& $0.34 \pm 0.01$ \\
German
& Subordination
& 2
& 0.62
& $0.91 \pm 0.00$
& $0.36 \pm 0.01$ \\
German
& Policy topic
& 13
& 0.08
& $0.71 \pm 0.03$
& $0.64 \pm 0.03$ \\
German
& Speech act
& 3
& 0.90
& $0.95 \pm 0.00$
& $0.17 \pm 0.01$ \\
\midrule
Italian
& Tense
& 3
& 0.71
& $0.92 \pm 0.01$
& $0.33 \pm 0.01$ \\
Italian
& Subordination
& 2
& 0.72
& $0.89 \pm 0.06$
& $0.23 \pm 0.10$ \\
Italian
& Policy topic
& 13
& 0.10
& $0.57 \pm 0.29$
& $0.49 \pm 0.30$ \\
Italian
& Speech act
& 3
& 0.96
& $0.97 \pm 0.01$
& $0.55 \pm 0.03$ \\
\midrule
Polish
& Tense
& 3
& 0.46
& $0.90 \pm 0.02$
& $0.50 \pm 0.02$ \\
Polish
& Subordination
& 2
& 0.73
& $0.91 \pm 0.01$
& $0.28 \pm 0.01$ \\
Polish
& Policy topic
& 13
& 0.08
& $0.78 \pm 0.03$
& $0.70 \pm 0.03$ \\
Polish
& Speech act
& 3
& 0.92
& $0.96 \pm 0.00$
& $0.14 \pm 0.01$ \\
\midrule
Turkish
& Tense
& 13
& 0.58
& $0.91 \pm 0.01$
& $0.66 \pm 0.00$ \\
Turkish
& Subordination
& 2
& 0.66
& $0.81 \pm 0.01$
& $0.26 \pm 0.00$ \\
Turkish
& Policy topic
& 13
& 0.08
& $0.73 \pm 0.03$
& $0.65 \pm 0.03$ \\
Turkish
& Speech act
& 3
& 0.93
& $0.96 \pm 0.01$
& $0.59 \pm 0.01$ \\
\bottomrule
\end{tabular}}
\caption{
Probe diagnostics by language and task. Majority is the proportion of
held-out examples belonging to the most frequent class. Accuracy is
held-out probe accuracy, and selectivity is held-out accuracy minus
accuracy under permuted labels
\citep{hewitt-liang-2019-designing}.
}
\label{tab:probe-diagnostics}
\end{table}

Table~\ref{tab:probe-diagnostics} shows why accuracy and selectivity
must be considered jointly. Tense selectivity ranges from $0.33$ to
$0.50$ in the four languages for which the task is available.
Policy-topic selectivity is also comparatively high
($0.49$--$0.70$), although the Italian result varies substantially
across backbones. Subordination selectivity ranges from $0.19$ to
$0.36$. Speech-act accuracy is uniformly high
($0.95$--$0.97$), but the majority class accounts for
$0.84$--$0.96$ of the held-out examples; accordingly, speech-act
selectivity is only $0.05$--$0.21$. We therefore treat these probes as
task-based operationalizations rather than exhaustive measurements of
morphology, syntax, semantics, and pragmatics, and interpret findings
based on lower-selectivity tasks with additional caution.

These diagnostics measure the predictability and class balance of the
automatically derived labels. They do not directly establish the
correctness of those labels across languages or historical periods.
A language- and period-stratified manual label audit remains necessary
to quantify label quality independently of probe performance.

\subsection{Additional trajectory and statistical details}
\label{app:trajectory-details}

\paragraph{Trajectory coordinates.}

For concept \(\kappa\), language \(\ell\), decade \(t\), and linguistic level
\(\lambda\), we average the activation vector over the shared features
assigned to that level:

\begin{equation}
    \mathbf{u}_{\lambda}(\kappa,\ell,t)
    =
    \frac{1}{|C_{\kappa,\ell,t}|}
    \sum_{s\in C_{\kappa,\ell,t}}
    \mathbf{f}_s[
        \mathcal{S}^{\mathrm{shared}}_{\lambda}
    ].
\end{equation}

A cell must contain at least 25 sentences. Cells with fewer than eight
available features are retained but marked as low coverage.

\paragraph{Alignment.}

The Convergence Index measures whether two languages become closer, but it
does not indicate whether their changes point in the same direction. We
therefore calculate the cosine between their net displacement vectors:

\begin{equation}
    \operatorname{par}_{\mathrm{end}}
    =
    \cos
    \left(
        \mathbf{u}_{\ell,t_{\mathrm{last}}}
        -
        \mathbf{u}_{\ell,t_{\mathrm{first}}},
        \mathbf{u}_{\ell',t_{\mathrm{last}}}
        -
        \mathbf{u}_{\ell',t_{\mathrm{first}}}
    \right).
\end{equation}

We also calculate stepwise alignment as the mean cosine between corresponding
decade-to-decade changes. A pair is labeled parallel when stepwise alignment
is above 0.20 and anti-parallel when it is below \(-0.20\). Values between
these thresholds remain unclassified.

\paragraph{Mechanism sharing.}

For every historical bin, we calculate the proportion of feature activation
mass assigned to cross-lingual shared features. We estimate its change over
time with a linear slope. A positive slope indicates increasing use of shared
features, while a flat or negative slope indicates that representational
convergence is not accompanied by greater mechanism sharing.

\paragraph{Mixed-effects specification.}

The main model is

\begin{equation}
    \operatorname{CI}
    \sim
    \operatorname{decade}
    \times
    \operatorname{concept\_group}
    \times
    \operatorname{level},
\end{equation}

with variance components for concept and language pair. Decade is centered
and scaled so that coefficients describe change per century. We fit the model
separately for every backbone. A pooled model includes backbone as an
additional variance component.

We report the direction of each effect, its variation across backbones, and
agreement on the resulting convergence category. Individual features are not
aligned across backbones because their feature spaces are learned
independently. Cross-model comparison is performed only at the level of the
estimated findings.

\section{Observable linguistic changes}
\label{app:observable-change}

The main analysis measures change in learned representations. To interpret
these aggregate changes, we also examine 18 observable linguistic measures
derived from the same sentence-level annotations used for feature attribution.
The measures are grouped into morphology, syntax, semantics, and pragmatics.
Examples include passive voice for morphology, subordination for syntax, human
subjects for semantics, and personal deixis for pragmatics.

We use the common 1950--2020 interval and balance the samples by decade,
sentence length, and topic. For each measure and language, we fit a linear
trend and report the estimated change over the full interval in percentage
points. A positive value means that the measure became more frequent between
1950 and 2020. A negative value means that it became less frequent.

We test whether each fitted trend differs from a flat trajectory and apply
Benjamini--Hochberg correction across the 90 language--measure tests. Bold
values in Table~\ref{tab:observable-measure-change} have $q<0.05$. Several
measures are based on terciles defined separately using each language's
historical distribution. Their absolute frequencies should therefore not be
compared across languages. The direction and size of their changes over time
can still be compared.

\begin{figure*}[t]
    \centering
    \includegraphics[width=\textwidth]{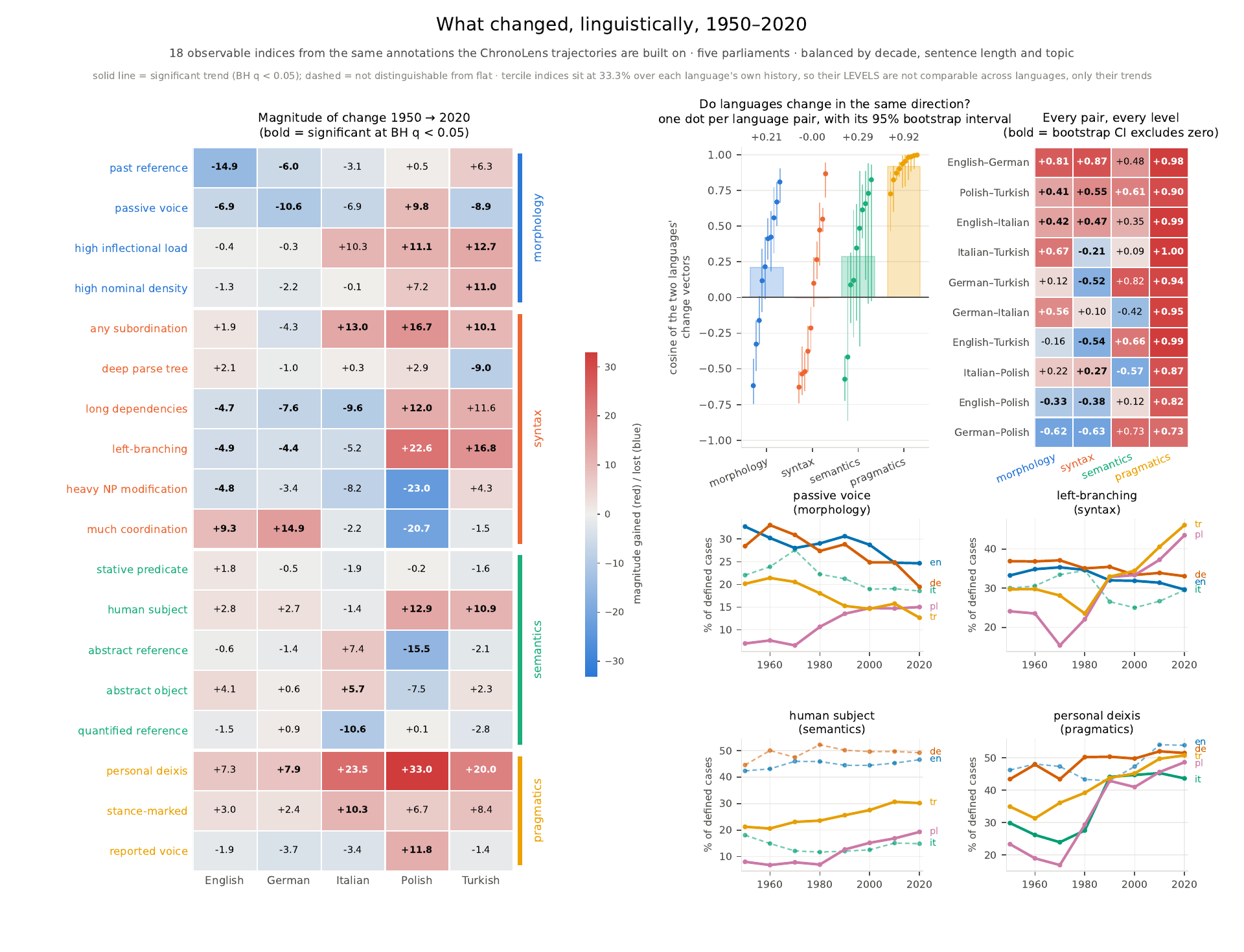}
    \caption{
    Observable linguistic changes between 1950 and 2020.
    The left panel reports the fitted change in 18 measures for five
    languages. Positive values indicate increases and negative values indicate
    decreases; bold values have Benjamini--Hochberg corrected $q<0.05$.
    The upper middle panel reports cosine similarity between each pair of
    languages, with 95\% bootstrap intervals. Positive cosine values indicate
    that the two languages tend to show increases and decreases in the same
    measures. Negative values indicate that measures that increase in one
    language tend to decrease in the other. Values near zero indicate no
    consistent shared direction. The upper right panel reports these
    similarities separately for morphology, syntax, semantics, and pragmatics.
    The lower panels show selected measure frequencies by decade. Tercile-based
    measures are defined separately for each language, so only their changes
    over time should be compared across languages.
    }
    \label{fig:observable-change}
\end{figure*}

\begin{table*}[t]
\centering
\scriptsize
\setlength{\tabcolsep}{4pt}
\resizebox{\textwidth}{!}{
\begin{tabular}{llrrrrr}
\toprule
\textbf{Level}
& \textbf{Measure}
& \textbf{English}
& \textbf{German}
& \textbf{Italian}
& \textbf{Polish}
& \textbf{Turkish} \\
\midrule

Morphology
& Past reference
& $\mathbf{-14.9}$
& $\mathbf{-6.0}$
& $-3.1$
& $+0.5$
& $+6.3$ \\

& Passive voice
& $\mathbf{-6.9}$
& $\mathbf{-10.6}$
& $-6.9$
& $\mathbf{+9.8}$
& $\mathbf{-8.9}$ \\

& High inflectional load
& $-0.4$
& $-0.3$
& $+10.3$
& $\mathbf{+11.1}$
& $\mathbf{+12.7}$ \\

& High nominal density
& $-1.3$
& $-2.2$
& $-0.1$
& $+7.2$
& $\mathbf{+11.0}$ \\

\midrule

Syntax
& Any subordination
& $+1.9$
& $-4.3$
& $\mathbf{+13.0}$
& $\mathbf{+16.7}$
& $\mathbf{+10.1}$ \\

& Deep parse tree
& $+2.1$
& $-1.0$
& $+0.3$
& $+2.9$
& $\mathbf{-9.0}$ \\

& Long dependencies
& $\mathbf{-4.7}$
& $\mathbf{-7.6}$
& $\mathbf{-9.6}$
& $\mathbf{+12.0}$
& $+11.6$ \\

& Left-branching
& $\mathbf{-4.9}$
& $\mathbf{-4.4}$
& $-5.2$
& $\mathbf{+22.6}$
& $\mathbf{+16.8}$ \\

& Heavy NP modification
& $\mathbf{-4.8}$
& $-3.4$
& $-8.2$
& $\mathbf{-23.0}$
& $+4.3$ \\

& Much coordination
& $\mathbf{+9.3}$
& $\mathbf{+14.9}$
& $-2.2$
& $\mathbf{-20.7}$
& $-1.5$ \\

\midrule

Semantics
& Stative predicate
& $+1.8$
& $-0.5$
& $-1.9$
& $-0.2$
& $-1.6$ \\

& Human subject
& $+2.8$
& $+2.7$
& $-1.4$
& $\mathbf{+12.9}$
& $\mathbf{+10.9}$ \\

& Abstract reference
& $-0.6$
& $-1.4$
& $+7.4$
& $\mathbf{-15.5}$
& $-2.1$ \\

& Abstract object
& $+4.1$
& $+0.6$
& $\mathbf{+5.7}$
& $-7.5$
& $+2.3$ \\

& Quantified reference
& $-1.5$
& $+0.9$
& $\mathbf{-10.6}$
& $+0.1$
& $-2.8$ \\

\midrule

Pragmatics
& Personal deixis
& $+7.3$
& $\mathbf{+7.9}$
& $\mathbf{+23.5}$
& $\mathbf{+33.0}$
& $\mathbf{+20.0}$ \\

& Stance-marked
& $+3.0$
& $+2.4$
& $\mathbf{+10.3}$
& $+6.7$
& $+8.4$ \\

& Reported voice
& $-1.9$
& $-3.7$
& $-3.4$
& $\mathbf{+11.8}$
& $-1.4$ \\

\bottomrule
\end{tabular}
}
\caption{
Signed fitted change in 18 observable linguistic measures between 1950 and
2020, measured in percentage points. Bold values have
Benjamini--Hochberg corrected $q<0.05$. Positive values indicate that a
measure became more frequent, while negative values indicate that it became
less frequent.
}
\label{tab:observable-measure-change}
\end{table*}

Of the 90 fitted trends, 37 remain significant after correction. Personal
deixis shows the clearest shared pattern: it increases in all five languages,
with significant increases in German, Italian, Polish, and Turkish. Several
other measures do not share one direction across languages. Long dependencies
decrease significantly in English, German, and Italian but increase
significantly in Polish. Left-branching decreases significantly in English and
German but increases significantly in Polish and Turkish. Passive voice
decreases significantly in English, German, and Turkish but increases
significantly in Polish. Coordination increases significantly in English and
German but decreases significantly in Polish. Thus, languages can have similar
overall amounts of change while differing in which linguistic properties
increase or decrease.

\subsection{Cross-language agreement}
\label{app:observable-agreement}

To measure whether two languages change in similar ways, we represent each
language by a vector containing its fitted changes for the measures within one
linguistic level. We then compute cosine similarity between the vectors of each
language pair. A positive value means that the two languages tend to increase
and decrease in the same measures. A negative value means that they tend to
change in different directions. A value near zero means that there is no
consistent relation between their patterns of change.

\begin{table*}[t]
\centering
\scriptsize
\setlength{\tabcolsep}{3.5pt}
\resizebox{\textwidth}{!}{
\begin{tabular}{lrrrrrrrr}
\toprule
\textbf{Level}
& \textbf{\# measures}
& \textbf{Mean cosine}
& \textbf{Minimum}
& \textbf{Maximum}
& \textbf{Pairs with CI excluding 0}
& \textbf{Leave-one-measure-out mean}
& \textbf{Sign agreement}
& \textbf{Timing $r$} \\
\midrule

Morphology
& 4
& $+0.21$
& $-0.62$ (DE--PL)
& $+0.81$ (EN--DE)
& $7/10$
& $[+0.08,+0.27]$
& $0.45$
& $+0.18$ \\

Syntax
& 6
& $0.00$
& $-0.63$ (DE--PL)
& $+0.87$ (EN--DE)
& $9/10$
& $[-0.07,+0.04]$
& $0.47$
& $+0.04$ \\

Semantics
& 5
& $+0.29$
& $-0.57$ (IT--PL)
& $+0.82$ (DE--TR)
& $3/10$
& $[+0.12,+0.35]$
& $0.56$
& $+0.07$ \\

Pragmatics
& 3
& $+0.92$
& $+0.73$ (DE--PL)
& $+1.00$ (IT--TR)
& $10/10$
& $[+0.53,+0.99]$
& $0.87$
& $+0.43$ \\

\bottomrule
\end{tabular}
}
\caption{
Cross-language agreement by linguistic level. Cosine similarity is computed
between the observable change vectors of each language pair. ``Pairs with CI
excluding 0'' reports how many of the ten pairwise 95\% bootstrap intervals
exclude zero. ``Leave-one-measure-out mean'' gives the range of the mean
cosine after removing one measure at a time. Sign agreement is the mean
proportion of measures with the same trend direction. Timing correlation is
the mean correlation between the corresponding decade-level trajectories.
}
\label{tab:observable-level-agreement}
\end{table*}

Table~\ref{tab:observable-level-agreement} shows that cross-language agreement
differs by linguistic level. Pragmatics has the strongest agreement, with a
mean cosine similarity of $0.92$. All ten language pairs have positive
bootstrap intervals that exclude zero. This result is driven mainly by the
widespread increase in personal deixis, although the leave-one-measure-out mean
remains positive, ranging from $0.53$ to $0.99$.

Morphology and semantics show weaker positive agreement, with mean cosine
similarities of $0.21$ and $0.29$. Syntax has a mean of $0.00$. This does not
mean that the syntactic measures remain unchanged. Instead, different language
pairs show different directions: four significant pairwise similarities are
positive and five are negative. There is therefore no single syntactic pattern
shared by all five languages.

\begin{table*}[t]
\centering
\scriptsize
\setlength{\tabcolsep}{5pt}
\begin{tabular}{lrrrrr}
\toprule
\textbf{Language pair}
& \textbf{Morphology}
& \textbf{Syntax}
& \textbf{Semantics}
& \textbf{Pragmatics}
& \textbf{All measures} \\
\midrule

English--German
& $\mathbf{+0.81}$
& $\mathbf{+0.87}$
& $+0.48$
& $\mathbf{+0.98}$
& $\mathbf{+0.81}$ \\

Polish--Turkish
& $\mathbf{+0.41}$
& $\mathbf{+0.55}$
& $\mathbf{+0.61}$
& $\mathbf{+0.90}$
& $\mathbf{+0.63}$ \\

English--Italian
& $\mathbf{+0.42}$
& $\mathbf{+0.47}$
& $+0.35$
& $\mathbf{+0.99}$
& $\mathbf{+0.51}$ \\

Italian--Turkish
& $\mathbf{+0.67}$
& $-0.21$
& $+0.09$
& $\mathbf{+1.00}$
& $\mathbf{+0.43}$ \\

German--Turkish
& $+0.12$
& $\mathbf{-0.52}$
& $\mathbf{+0.82}$
& $\mathbf{+0.94}$
& $+0.01$ \\

German--Italian
& $\mathbf{+0.56}$
& $+0.10$
& $-0.42$
& $\mathbf{+0.95}$
& $\mathbf{+0.36}$ \\

English--Turkish
& $-0.16$
& $\mathbf{-0.54}$
& $\mathbf{+0.66}$
& $\mathbf{+0.99}$
& $-0.01$ \\

Italian--Polish
& $+0.22$
& $\mathbf{+0.27}$
& $\mathbf{-0.57}$
& $\mathbf{+0.87}$
& $\mathbf{+0.38}$ \\

English--Polish
& $\mathbf{-0.33}$
& $\mathbf{-0.38}$
& $+0.12$
& $\mathbf{+0.82}$
& $-0.03$ \\

German--Polish
& $\mathbf{-0.62}$
& $\mathbf{-0.63}$
& $\mathbf{+0.73}$
& $\mathbf{+0.73}$
& $\mathbf{-0.23}$ \\

\bottomrule
\end{tabular}
\caption{
Cosine similarity between observable change vectors for every language pair
and linguistic level. Bold values have 95\% bootstrap intervals that exclude
zero. The final column computes cosine similarity over all 18 measures.
}
\label{tab:observable-pairwise-agreement}
\end{table*}

Table~\ref{tab:observable-pairwise-agreement} shows that the direction of
change depends on both the language pair and the linguistic level. For example,
English and German have positive similarities for morphology, syntax, and
pragmatics, while German and Polish have negative similarities for morphology
and syntax but positive similarities for semantics and pragmatics. A positive
overall value therefore does not imply agreement at every linguistic level.

The pairwise results are not explained by language-family membership alone.
The mean cosines for pairs within the Indo-European group and pairs involving
Turkish are, respectively, $0.18$ and $0.26$ for morphology, $0.12$ and
$-0.18$ for syntax, $0.12$ and $0.55$ for semantics, and $0.89$ and $0.95$
for pragmatics. Some pairs involving Turkish have high semantic or pragmatic
similarity, while some pairs of Indo-European languages have negative
similarity for morphology or syntax. Because the analysis contains only five
languages, these comparisons are descriptive. They do not establish a general
relation between genealogical relatedness and the direction of historical
change.

\subsection{Long-window results}
\label{app:observable-long-window}

The common 1950--2020 interval supports direct comparison across languages but
does not use the full historical record available for English, German, Italian,
or Polish. Table~\ref{tab:observable-long-window} therefore reports the largest
significant trends over each language's complete available interval. The
values are fitted percentage-point changes per century. They should not be
compared as total changes because the historical spans differ.

\begin{table*}[t]
\centering
\small
\begin{tabularx}{\textwidth}{@{}llX@{}}
\toprule
\textbf{Language}
& \textbf{Span}
& \textbf{Largest significant changes per century} \\
\midrule

English
& 1800--2020
& Past reference $-34.5$; personal deixis $+21.5$; human subject
$+15.0$; passive voice $-12.1$. \\

German
& 1860--2020
& Personal deixis $+22.1$; human subject $+20.7$; passive voice
$-11.3$; heavy NP modification $-8.3$. \\

Italian
& 1840--2020
& Coordination $+11.7$; left-branching $-9.3$; heavy NP modification
$+8.8$; human subject $-8.4$. \\

Polish
& 1910--2020
& High nominal density $+11.6$; deep parse tree $+9.2$. \\

Turkish
& 1950--2020
& Personal deixis $+28.8$; left-branching $+24.3$; high inflectional
load $+18.3$; high nominal density $+15.8$. \\

\bottomrule
\end{tabularx}
\caption{
Largest significant observable changes over each language's complete
historical record. Values are fitted percentage-point changes per century, and
all listed trends have $q<0.05$. The unequal intervals make this analysis
supplementary to the balanced 1950--2020 comparison.
}
\label{tab:observable-long-window}
\end{table*}

\end{document}